\documentclass[11pt,a4paper]{article}
\usepackage{times,latexsym}
\usepackage{url}
\usepackage[T1]{fontenc}
\usepackage{tabularx}
\usepackage{xcolor}
\usepackage{graphics}
\usepackage{microtype}
\usepackage{inconsolata}
\usepackage{amsmath}
\usepackage{booktabs}
\usepackage{float}
\usepackage{multirow}
\usepackage{hyperref}

\usepackage{algorithm}
\usepackage{algorithmic}

\usepackage{xcolor}
\definecolor{blue}{HTML}{a1c9f4}
\definecolor{orange}{HTML}{ffb482}
\definecolor{green}{HTML}{8de5a1}
\definecolor{red}{HTML}{ff9f9b}
\definecolor{purple}{HTML}{d0bbff}
\definecolor{brown}{HTML}{debb9b}
\definecolor{pink}{HTML}{fab0e4}
\definecolor{grey}{HTML}{cfcfcf}
\definecolor{lightyellow}{HTML}{fffea3}
\definecolor{paleblue}{HTML}{b9f2f0}

\definecolor{lgrey}{HTML}{787878}
\definecolor{mgrey}{HTML}{656565}
\definecolor{black}{HTML}{000000}
\usepackage{amsmath}
\usepackage{amssymb}
\usepackage{mathtools}
\usepackage{amsthm}
\usepackage{soul}

\usepackage{inconsolata}
\usepackage{float}
\usepackage{multirow}
\usepackage{bm}
\usepackage{bbm}
\usepackage{enumitem}
\theoremstyle{plain}

\theoremstyle{definition}

\theoremstyle{remark}

\usepackage{xspace}
\newcommand{\method}{{\textsc{AutoPEFT}}\xspace}
\usepackage{todonotes}

\usepackage[acceptedWithA]{tacl2021v1}

\usepackage{xspace,mfirstuc,tabulary}

\newif\iftaclinstructions
\taclinstructionsfalse %
\iftaclinstructions
\renewcommand{\confidential}{}
\renewcommand{\anonsubtext}{(No author info supplied here, for consistency with
TACL-submission anonymization requirements)}
\newcommand{\instr}
\fi

\iftaclpubformat %

\else

\fi

\title{\method: Automatic Configuration Search for \\ Parameter-Efficient Fine-Tuning}

\author{Han Zhou\textsuperscript{1,*}
\quad
Xingchen Wan\textsuperscript{2,*}
\quad
Ivan Vuli{\'c}\textsuperscript{1}
\quad
Anna Korhonen\textsuperscript{1}
 \\
  \textsuperscript{1}Language Technology Lab, University of Cambridge \\
  \textsuperscript{2}Machine Learning Research Group, University of Oxford \\
  \texttt{\{hz416, iv250, alk23\}@cam.ac.uk} \\
  \texttt{xwan@robots.ox.ac.uk}
}

\date{}

\begin{document}
\maketitle
\begin{abstract}
Large pretrained language models are widely used in downstream NLP tasks via task-specific fine-tuning, but such procedures can be costly. Recently, Parameter-Efficient Fine-Tuning (PEFT) methods have achieved strong task performance while updating much fewer parameters than full model fine-tuning (FFT). However, it is non-trivial to make informed design choices on the \textit{PEFT configurations}, such as their architecture, the number of tunable parameters, and even the layers in which the PEFT modules are inserted. Consequently, it is highly likely that the current, manually designed configurations are suboptimal in terms of their performance-efficiency trade-off. Inspired by advances in neural architecture search, we propose \method for automatic PEFT configuration selection: we first design an expressive configuration search space with multiple representative PEFT modules as building blocks. Using multi-objective Bayesian optimisation in a low-cost setup, we then discover a Pareto-optimal \textit{set} of configurations with strong performance-cost trade-offs across different numbers of parameters that are also highly transferable across different tasks. Empirically, on GLUE and SuperGLUE tasks, we show that \method-discovered configurations significantly outperform existing PEFT methods and are on par or better than FFT without incurring substantial training efficiency costs.
\end{abstract}
\section{Introduction and Motivation}
\label{sec:introduction}
\def\thefootnote{*}\footnotetext{Equal contribution.}Pretrained language models (PLMs) are used in downstream tasks via the standard transfer learning paradigm, where they get fine-tuned for particular tasks~\cite{devlin-etal-2019-bert, DBLP:journals/corr/abs-1907-11692}. This achieves state-of-the-art results in a wide spectrum of NLP tasks, becoming a prevalent modelling paradigm in NLP~\cite{DBLP:journals/jmlr/RaffelSRLNMZLL20}. Fine-tuning the PLMs typically requires a full update of their original parameters (i.e. the so-called \textit{full-model fine-tuning (FFT)}); however, this is (i) computationally expensive and also (ii) storage-wise expensive as it requires saving a separate full model copy for each task-tuned model. With the ever-growing size of the PLMs~\cite{DBLP:conf/nips/BrownMRSKDNSSAA20, DBLP:conf/iclr/SanhWRBSACSRDBX22}, the cost of full-model FT becomes a major bottleneck, due to its increasing demands as well as computational (time and space) non-efficiency.

\begin{figure}[!t]
    \centering
    \includegraphics[width=0.95\linewidth, trim={0cm 0 0 0},clip]{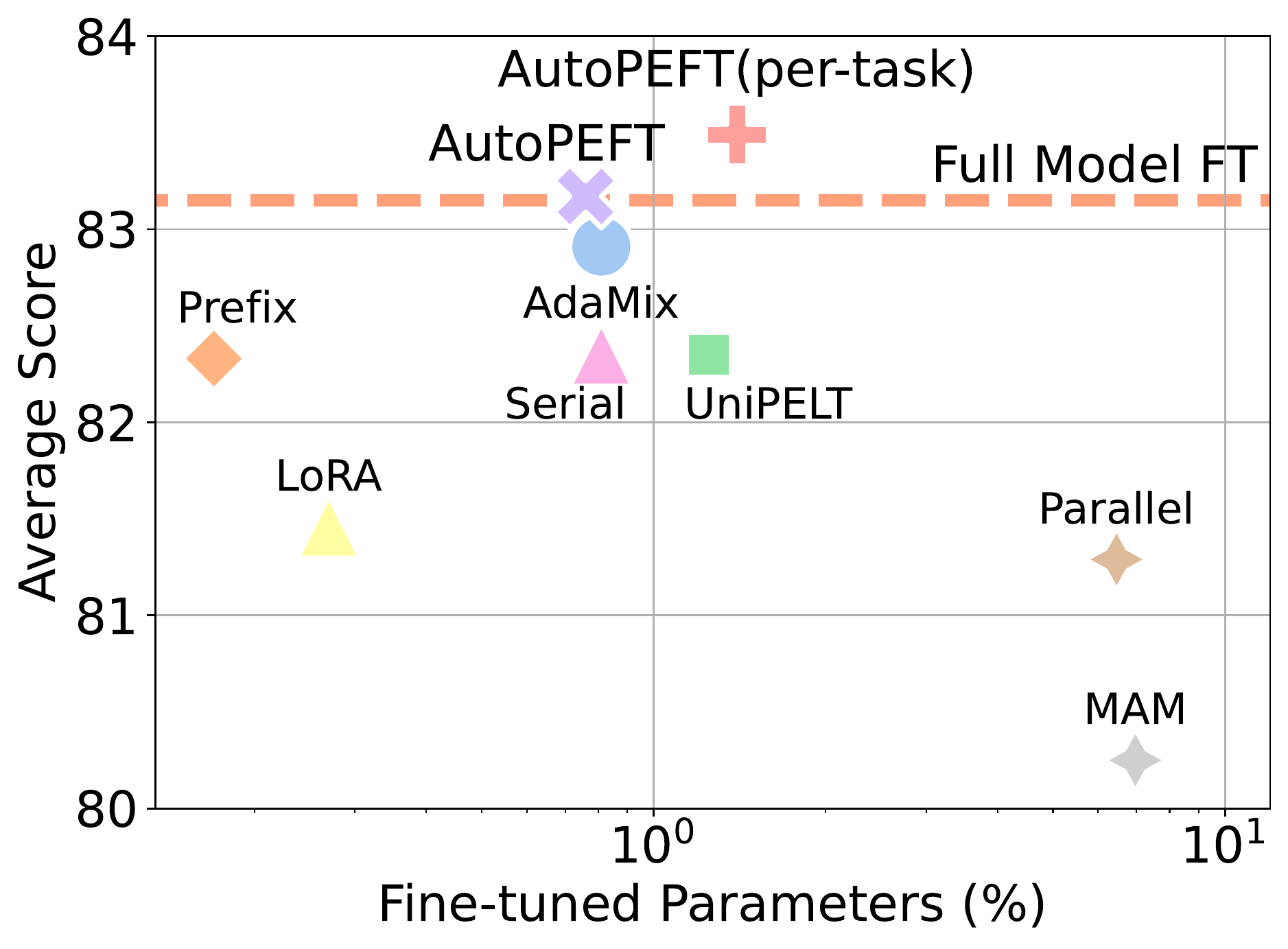}
    \vspace{-2mm}
    \caption{Performance of \method-discovered configurations (\texttt{AutoPEFT} \& \texttt{AutoPEFT(per-task)}; see details in Table \ref{tab:keyresults}) compared to other baseline PEFT methods (markers) and full model FT that updates 100\% of parameters (dashed horizontal bar), averaged across 8 GLUE tasks. Our approach achieves the best trade-off between task performance and parameter efficiency.}
    \vspace{-2mm}
    \label{fig:my_label}
\end{figure}

Parameter-efficient fine-tuning (PEFT) delivers a solution for alleviating the issues with full-model FT~\cite{DBLP:conf/icml/HoulsbyGJMLGAG19}. By freezing the majority of pretrained weights of PLMs, PEFT approaches only update a small portion of parameters for efficiently adapting the PLM to a new downstream task. Recent studies have shown that PEFT can achieve competitive task performance while being modular, adaptable, and preventing catastrophic forgetting in comparison to traditional FFT~\cite{https://doi.org/10.48550/arxiv.2205.12410,Pfeiffer:2023survey}. 

Recent developments have created diverse PEFT modules with distinctive characteristics~\cite{pfeiffer-etal-2020-mad, li-liang-2021-prefix}, with one of the two main aims in focus: \textbf{1)} \textit{improve task performance} over other PEFT approaches while \textit{maintaining the same parameter budget} as the competitor PEFT methods; or \textbf{2)} \textit{maintain task performance} while \textit{reducing the parameter budget} needed. Existing PEFT modules, optimising for one of the two aims, have been successfully applied to transfer learning tasks~\cite{https://doi.org/10.48550/arxiv.2205.13535, pfeiffer-etal-2022-lifting}.  However, different tasks, with different complexity, show distinct sensitivity to the allocated parameter budget and even to the chosen PEFT approach~\cite{DBLP:conf/iclr/HeZMBN22}. At the same time, most PEFT applications are limited to a single PEFT architecture (e.g. serial adapters, prefix-tuning) with fixed decisions on its components (e.g. hidden size dimensionality, insertion layers) resulting in \textit{potentially suboptimal PEFT configurations} across many tasks. Therefore, in this work, we propose a new, versatile and unified framework that automatically searches for improved and task-adapted PEFT configurations, aiming to \textit{effectively balance} between the two (often colliding goals) of (i) improving performance and (ii) keeping the desired low parameter budget for PEFT.

While recent research has started exploring more dynamic PEFT configurations, the prior studies remain limited across several dimensions, including how they define the configuration search space. Namely, they typically focus only on a single PEFT architecture (e.g. adapters) or their simple combinations, or a single property (e.g. insertion layers -- where to insert the module); see a short overview later in \S\ref{sec:rw}. Here, we propose a unified and more comprehensive framework for improved configuration search. It covers multiple standard PEFT modules (serial adapters, parallel adapters, and prefix-tuning) as building blocks, combined with the critical parameter budget-related decisions: the size of each constituent module and the insertion layers for the modules.

Our defined comprehensive search space is huge; consequently, traversing it effectively \textit{and} efficiently is extremely challenging. To enable search over the large configuration space, we thus propose the novel \method framework. It \underline{auto}matically configures multiple \underline{PEFT} modules along with their efficiency-oriented design decisions, relying on a high-dimensional Bayesian optimisation (BO) approach. Crucially, within the search space, we propose a multi-objective optimisation which learns to balance simultaneously between maximising the searched configurations' task performance \textit{and} parameter efficiency.

We conduct extensive experiments on the standard GLUE and SuperGLUE benchmarks~\cite{wang-etal-2018-glue, DBLP:conf/nips/WangPNSMHLB19}, with encoder-only and encoder-decoder models. We first study the transferability of the \method-searched architecture by running \method on a single task with a low-fidelity proxy (aiming to reduce computational cost), followed by transferring the found architecture to other tasks.  Experimental results show that this architecture can outperform existing PEFT baselines while achieving on-par performance with the standard FFT. Further slight gains can be achieved with a larger computation budget for training, where we run \method per task to find a task-adapted PEFT configuration. As revealed in Figure~\ref{fig:my_label}, \method can find configurations that offer a solid trade-off between task performance and parameter efficiency, even outperforming FFT. We also provide ablation studies over the search space, validating that the \method framework is versatile and portable to different search spaces.

\vspace{1mm}
\noindent \textbf{Contributions.} 
\textbf{1)} We propose the \method search space containing diverse and expressive combinations of PEFT configurations from three representative PEFT modules as foundational building blocks and the binary decisions concerning Transformer layers for inserting these modules as searchable dimensions. \textbf{2)} To navigate the vast \method search space and to discover {a 
\emph{set} of} transferable PEFT configurations that optimally trade performance against cost across various parameter ranges \emph{in a single run}, we further propose an effective search method based on multi-dimensional Bayesian optimisation. \textbf{3)} We demonstrate that the one-time search cost of \method is low, and \method yields task-shareable configurations, outperforming existing PEFT modules while being transferable across tasks. The \method framework can also be easily extended to other and new PEFT modules. The code is available at \url{https://github.com/cambridgeltl/autopeft}.

\begin{figure*}[ht]
  
    \centering
    \includegraphics[width=0.97\linewidth]{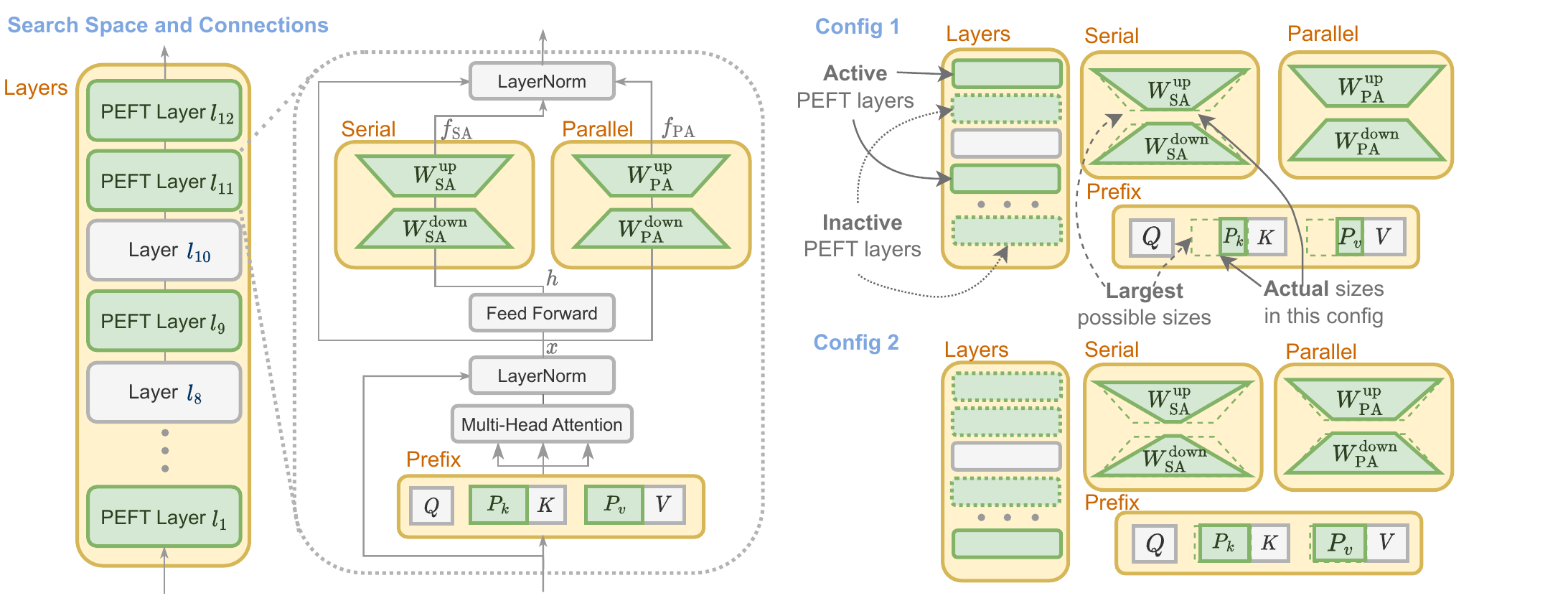}
    \vspace{-3mm}
    \caption{Illustration of the \method \emph{search space} which combines both layer-level (\texttt{Layers}) and within-layer (\texttt{Serial, Parallel, Prefix}) search, and the connections within a layer  (\textbf{Left}). We further show two possible \emph{configurations} in the search space (\textbf{Right}): note that some PEFT layers can be inactive altogether and the searchable module sizes (shaded in \textcolor{green}{green}), i.e. the bottleneck sizes in \texttt{Serial} and \texttt{Parallel} ($D_{\text{SA}}$ and $D_{\text{PA}}$ respectively) and sizes of $P_K, P_V$ in \texttt{Prefix} ($L_{\text{PT}}$), are dynamic.}
        \vspace{-1mm}
    \label{fig:architecture and search space}

\end{figure*}

\section{\method Framework}
\label{sec:autopeft}

\subsection{Designing the \method Search Space}
\label{sec:space}

Inspired by the success of neural architecture search (NAS) methodology \cite{ru2020neural}, we similarly start by designing a large and expressive configuration space. We additionally provide the motivation behind each decision to include a particular module and its components in the configuration space, along with a mathematical formulation.

\label{sec:search_space}
The search space is known to be one of the most important factors in the performance of the configurations to be discovered subsequently \cite{ru2020neural, xie2019exploring, li2020random, dong2020bench, yang2019evaluation}. In order to simultaneously maximise task performance along with parameter efficiency, it is necessary to first define a `parameter-reducible' search space, where each dimension within the space potentially contributes to reducing the parameter budget. Similarly, each dimension potentially impacts the performance positively without introducing redundancy in the space~\cite{DBLP:conf/iclr/WanREL22}. Therefore, we propose the following search space with representative PEFT modules spanning a plethora of (non-redundant) configurations as illustrated in Figure~\ref{fig:architecture and search space}:

\vspace{0.8mm}
\noindent \textit{PEFT Modules.} Inspired by common practices in NAS of using known well-performing modules as building blocks, we include three distinctive PEFT designs to efficiently adapt different forwarding stages of hidden states in the PLM layers. We combine Serial Adapters (SA), Parallel Adapters (PA), and Prefix-Tuning (PT) as the three representative modules in the search space as the building blocks, where the PT module adapts the multi-head attention layer, and SA and PA interact with the FFN layer (Figure~\ref{fig:architecture and search space}). Each configuration makes a decision on the PEFT modules in the insertion layer: all of them can be `turned' on or off. We combine this binary decision with the actual non-binary decision on the module size (see next) so that the value of $0$, in fact, denotes the absence of the modules in the layer(s). We note that other PEFT modules such as LoRA \citep{DBLP:conf/iclr/HuSWALWWC22} are scaled variants of PA with the same insertion form \citep{DBLP:conf/iclr/HeZMBN22}. As we empirically validate later, the resultant search space spanned by the selected building blocks is extremely expressive and flexible and enables the discovery of configurations that outscore any of the individual building blocks and other PEFT modules.

\vspace{0.8mm}
\noindent \textit{Size.} 
Previous studies show that PEFT methods are highly sensitive to the number of tunable parameters: adaptively setting their capacity in accordance with the target task is, therefore, essential for achieving good performance~\cite{Chen:2022emnlp}. The number of tunable parameters depends on each particular module. The additional parameters introduced by both SA and PA are dominated by their bottleneck dimension $D$. Similarly, the size of the PT module is defined by its prefix length $L_{\mathrm{PT}}$. Thus, we define a binary logarithmic search scale for the respective discrete sets $D_{\mathrm{SA}}$, $D_{\mathrm{PA}}$, and $L_{\mathrm{PT}}$, spanning the values from 0 (absence of the module) to $D_{\mathrm{h}}$ where $D_{\mathrm{h}}$ is the dimensionality of the output embedding of the PLM (e.g. $D_{\mathrm{h}}$=$768$ for BERT\textsubscript{base}).

\vspace{0.8mm}
\noindent \textit{Insertion Layers.} Prior work has also shown that different layers in the PLMs store different semantic information~\cite{vulic-etal-2020-probing}, where the higher layers produce more task-specific and contextualized representations~\cite{tenney-etal-2019-bert}. Therefore, as another configuration dimension, we aim to search for the minimal number and the actual position of layers in which to insert the PEFT modules. We define a binary `insertion' decision at each layer $l_{i}$.

\begin{figure}[!t]
    \centering
    \includegraphics[width=0.97\linewidth,trim={0 0 0.2cm 0},clip]{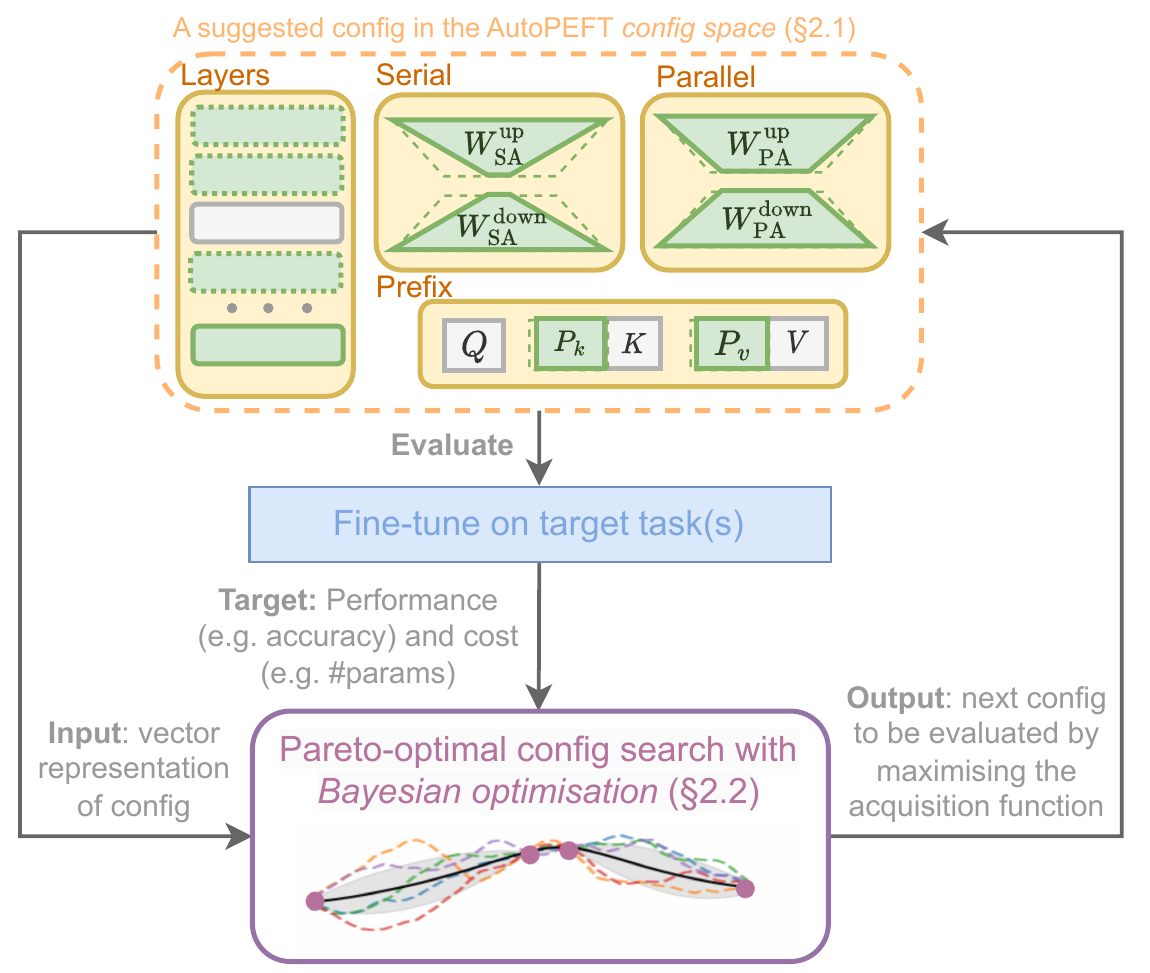}
    \vspace{-2mm}
    \caption{{Illustration of the Pareto-optimal search with multi-objective Bayesian optimisation (BO; §\ref{sec:bo_method}): The BO agent trains on the vector representations of the evaluated configurations as inputs and their performance under a low-fidelity setup (e.g. accuracy -- obtained by fine-tuning the language model with the PEFT configuration for a small number of iterations) and cost (e.g. number of parameters) as targets. The BO agent then iteratively suggests new configurations until convergence.}}
    \label{fig:demonstration}
    \vspace{-1.5mm}
\end{figure}

\vspace{1.3mm}
\noindent \textbf{Combining PEFT Modules.}
The SA module and the PA module share a bottleneck architecture. The SA receives hidden states from the FFN output as its inputs, adapting it with a down-projection matrix $W^{\mathrm{down}}_{\mathrm{SA}}\in\mathbb{R}^{D_{\mathrm{h}}\times D_{\mathrm{SA}}}$, followed by a non-linear activation function, and then an up-projection matrix $W^{\mathrm{up}}_{\mathrm{SA}}\in\mathbb{R}^{D_{\mathrm{SA}}\times D_{\mathrm{h}}}$:
\begin{equation}
    f_{\mathrm{SA}}(h) = \mathrm{ReLU}(hW^{\mathrm{down}}_{\mathrm{SA}})W^{\mathrm{up}}_{\mathrm{SA}}.
\end{equation}
PA, on the other hand, receives its inputs from hidden states before the FFN layer with the same formulation:
\begin{equation}
    f_{\mathrm{PA}}(x) = \mathrm{ReLU}(xW^{\mathrm{down}}_{\mathrm{PA}})W^{\mathrm{up}}_{\mathrm{PA}}.
\end{equation}
Therefore, it is able to act in parallel with the SA without interference. Note that the FFN hidden states $h=F(x)$ contain the task-specific bias learned in its pretrained weights. Therefore, by combining SA with PA, the following composition of functions is achieved:
\begin{equation}
\begin{split}
f_{\mathrm{SAPA}}(x) = &\mathrm{ReLU}(F(x)W^{\mathrm{down}}_{\mathrm{SA}})W^{\mathrm{up}}_{\mathrm{SA}} \\
+ &\mathrm{ReLU}(xW^{\mathrm{down}}_{\mathrm{PA}})W^{\mathrm{up}}_{\mathrm{PA}}.
\end{split}
\label{eq:sapa}
\end{equation}

The final composition should adapt effectively to both bias-influence hidden states and the original inputs before the pretrained FFN layer.\footnote{The PA module also acts as the low-rank reparameterisation of the learned SA and the frozen FFN layer to further match the intrinsic dimensionality of the target task.}

Further, applying PEFT modules to interact with FFNs and multi-head attention should positively impact task performance~\cite{mao-etal-2022-unipelt, DBLP:conf/iclr/HeZMBN22}. PT learns two prefix vectors, $P_k$ and $P_v \in \mathbb{R}^{L_{\mathrm{PT}}\times D_{\mathrm{h}}}$, that are concatenated with the original multi-head attention's key and value vectors, which efficiently adapts the multi-head attention layer to fit the target task. Thus, we finally combine the SA and the PA (i.e., SAPA from above) with PT.

In sum, the overview of the dimensions spanning the final configuration space is provided in Figure~\ref{fig:architecture and search space}. The combination of the different `configuration dimensions' outlined above gives rise to a total of, e.g. 5,451,776 possible configurations with BERT\textsubscript{base} and $\sim3\times$$10^{10}$ configurations with RoBERTa\textsubscript{large} (i.e. the number of configurations is $2^{|l|} \times |D_\mathrm{SA}|\times|D_\mathrm{PA}|\times|L_\mathrm{PT}|$). While a large search space is crucial for expressiveness and to ensure that good-performing configurations are contained, it also increases the difficulty for search strategies to navigate the search space well while remaining sample- and thus computationally efficient. Furthermore, in the PEFT setting, we are also often interested in discovering a family of configurations that trade-off between performance and efficiency for general application in various scenarios with different resource constraints, thus giving rise to a multi-objective optimisation problem where we simultaneously aim to maximise performance while minimising costs. In what follows, we propose a search framework that satisfies all those criteria.

\subsection{Pareto-Optimal Configuration Search}
\label{sec:bo_method}

\begin{algorithm}[h]
\caption{Overall \method search pipeline.}\label{alg:autopeft}
\small
\begin{algorithmic}[1]
    \STATE {\bfseries Input:} number of randomly initialising points $N_0$, maximum number of config evaluations $N > N_0$, \method search space $\mathcal{A}$.
    \STATE {\bfseries Output:} a \textit{set} of Pareto-optimal configs $A^*$.
    \STATE Initialise by sampling randomly at $N_0$ configurations $a \sim \mathcal{A}$ and fine-tune the PLM to obtain $\bm{f}(\cdot)$ of the corresponding configs. Initialise $\mathcal{D}_0 \leftarrow \{(a_i, \bm{f}(a_i))\}_{i=1}^{N_0}$ and fit a SAAS-GP model on $\mathcal{D}_0$.
    \FOR{$n=N_0, \dots, N$}
            \STATE Select the next configuration(s) to evaluate $a_n$ by maximising the NEHVI acquisition function $a_n = \mathrm{argmax}_{a \in \mathcal{A}}\alpha (a|\mathcal{D}_{n-1})$.
            \STATE Fine-tune the PLM with candidate configuration(s) $a$ (possibly with low-fidelity estimates) to obtain $\bm{f}(a)$ \textcolor{gray}{// \textit{Inner-loop optimisation in Eq.~\ref{eq:bo_objective}}}.
            \STATE Augment the observation data $\mathcal{D}_n \leftarrow \mathcal{D}_{n-1}  \cup (a_t, \bm{f}(a_t) )$ and update the SAAS-GP model.
    \ENDFOR
    \STATE Return the set of non-dominated configurations $A^* \subseteq \{a_i\}_{i=1}^{N}$.
\end{algorithmic}
\end{algorithm}

\vspace{1.2mm}
\noindent \textbf{Multi-objective Optimisation Formulation.} The ultimate goal of \method is to discover promising PEFT configuration(s) from the expressive search space designed in §\ref{sec:search_space}, which is itself challenging. In this paper, we focus on an even more challenging but practical goal: instead of aiming to find a {single, best-performing} PEFT configuration, we aim to discover \emph{a family of Pareto-optimal} PEFT configurations that trade performance against parameter-efficiency (or parameter cost) optimally: one of the most impactful use cases of PEFT is its ability to allow fine-tuning of massive language models even with modest computational resources, and thus we argue that searching Pareto-optimal configurations is key as it allows tailored user- and scenario-specific PEFT deployment depending on the computational budget.

Formally, denoting the full \method search space as $\mathcal{A}$ and a single configuration $a \in \mathcal{A}$ with trainable weights $W$, without loss of generality, assuming our objective is to maximise (i) a performance metric $f(a, W)$ (e.g. the accuracy on the dev set) and to (ii) minimise a cost metric $g(a)$ (e.g. the number of parameters in $a$), a search method aims to solve the bi-level, bi-objective optimisation problem:
\begin{align}
    \begin{split}
    & \max_{a \in \mathcal{A}}\Big( f(a, W^*(a)), -g(a) \Big);\\
    \mathrm{s.t. } & W^*(a) = \arg \min_{W} \mathcal{L}_{\mathrm{train}}(a, W),
\end{split}
    \label{eq:bo_objective}
\end{align}
where the inner loop optimisation problem is the optimisation of the configuration \textit{weights} achieved by fine-tuning the configuration $a$ itself over the training loss $\mathcal{L}_{\mathrm{train}}$. Given the bi-objective nature of the problem, there is, in general, no single maximiser of Eq.~\eqref{eq:bo_objective} but a \textit{set} of Pareto-optimal configurations $A^* = \{a^*_1, ..., a^*_{|A^*|} \}$.
that are \textit{non-dominated}: we say that a configuration $a$ dominates another $a'$ (denoted $\bm{f}(a') \prec \bm{f}(a)$) if $\mathcal{L}_{\mathrm{val}}(a, W^*) \leq \mathcal{L}_{\mathrm{val}}(a', W^*)$ \emph{and} $g(a)\leq g(a')$ and either $\mathcal{L}_{\mathrm{val}}(a, W^*) < \mathcal{L}_{\mathrm{val}}(a', W^*)$ or $g(a)< g(a')$. Denoting $\bm{f}(a) := [\mathcal{L}_{\mathrm{val}}(a, W^*), g(a)]^{\top}$, the set of Pareto-optimal architectures $A^*$ are those that are mutually non-dominated: $A^* = \{a^*_i \in \mathcal A ~|~  \nexists \text{ }  a' \in \mathcal{A} \text{ s.t. } \bm{f}(a') \prec \bm{f}(a^*_i) \}$. The Pareto front (PF) $\mathcal P^*$ is the image of the Pareto set of architectures: $\mathcal {P}^* = \{\bm{f}(a) ~|~a\in A^*\}$.

\vspace{1.2mm}
\noindent \textbf{Bayesian Optimisation (BO).} To solve Eq.~\eqref{eq:bo_objective}, we adopt a BO approach, illustrated in Figure~\ref{fig:demonstration}. On a high level, BO consists of a \textit{surrogate model} that sequentially approximates the objective function based on the observations so far and an \textit{acquisition function} that is optimised at each iteration to actively select the next configuration to evaluate. Typically, the surrogate model is a Gaussian process (GP), a flexible and non-parametric model with well-principled and closed-form uncertainty estimates: given an observed set of $n$ configurations and their evaluated performance: $\mathcal{D}_n = \{ \big( a_i, \bm{f}(a_i) \big) \}_{i=1}^n$, the GP surrogate model gives a closed form posterior distribution $\mathbb{P}(\bm{f}(a)| \mathcal{D}_n)$ over the true, unobserved function values $\bm{f}$ potentially over configurations that have \textit{not} been evaluated before. The acquisition function $\alpha: \mathcal{A} \rightarrow \mathbb{R}$, on the other hand, uses the posterior distribution of the surrogate model to assign a utility value to possible configuration candidates in $\mathcal{A}$, typically balancing exploitation (i.e., querying near configurations in $\{a_i\}_{i=1}^n$ that were previously observed to be strong) and exploration (i.e., the configurations far from $\{a_i\}_{i=1}^n$ and are those we do not have knowledge on and can potentially be even better configurations). At each step of BO, the acquisition function is optimised (note that while evaluating $\bm{f}(a)$ is expensive, evaluating $\alpha(a|\mathcal{D})$, which only uses the posterior distribution from the surrogate model, is not) to select the next configuration (or batch of configurations) $a_{n+1} = \arg\max_{a \in \mathcal{A}} \alpha(a|\mathcal{D}_n)$ to evaluate. For a detailed overview of BO, we refer the readers to~\citet{garnett2023bayesian} and \citet{ frazier2018tutorial}. 

\vspace{1.2mm}
\noindent \textbf{Rationales for Using BO.} We argue that BO is well-suited to the task in principle and has various advantages over alternative, viable approaches such as those based on differentiable NAS (DARTS) \cite{liu2018darts}, which typically utilise a \textit{continuous relaxation} of the discrete configurations, thereby allowing $a$ to be \textit{jointly} optimised with the model weights $W$ in Eq.~\ref{eq:bo_objective} with a supernet:

First, unlike the DARTS-based approach, by treating the optimisation problem defined in Eq.~\ref{eq:bo_objective} as a \textit{black box}, BO decouples the optimisation of the weights $W$ and the optimisation of architecture $a$, and solves the latter problem with no gradient information at all \cite{white2021bananas, ru2020interpretable}. This makes a BO-based solution more parallelisable and more amenable to a distributed setup, which modern large PLMs often rely on, as multiple configuration evaluations may take place simultaneously in different client machines as long as they can relay the evaluation results $\bm{f}$ back to a central server running the BO. This further contributes to memory efficiency, as unlike the DARTS-based method that optimises a supernet (a heavily over-parameterised network that can be deemed as a weighted superposition of all configurations in $\mathcal{A}$), each parallel evaluation in BO trains a single configuration only; we argue that this point is particularly important for PEFT given its main promise on \textit{parameter efficiency}. 

Second, as discussed, it is often desirable to discover a \textit{family} of configurations with different trade-offs between performance and parameters in different application scenarios. As we will show, while BO generalises elegantly to handle vector-valued objective functions and may generate a PF of configurations \textit{in a single run}, competing methods, such as supernet-based NAS methods, typically require a scalar objective function and thus are limited to discovering a single best-performing configuration~\cite{eriksson2021latency, guerrero2021bag}; this means that one typically needs to run the NAS pipeline multiple times for different cost budgets in these methods. 

Lastly, while one of the main arguments favouring differentiable techniques is its lighter computational expense as one only needs to train the supernet once rather than repeatedly training different candidate configurations, as we will later show, the sample-efficient nature of BO and strong transferability of the discovered configurations also ensure that the computational cost of our proposed method remains tractable. As we will show in \S\ref{sec:experiments}, while DARTS-based NAS is indeed a plausible approach for PEFT configuration search, we show that our approach performs competitively to S\textsuperscript{3}PET \cite{DBLP:journals/corr/abs-2206-07382}, a DARTS-based method.

\vspace{1.2mm}
\noindent \textbf{Adapting BO to the \method Task.} Adapting BO to the high-dimensional and combinatorial \method search space is non-trivial.
To address the challenges, we customise both components of BO, and the overall pipeline is shown in Algorithm~\ref{alg:autopeft}. Instead of a standard GP, we propose to use a \textit{Gaussian process with sparse axis-aligned subspaces} (SAAS-GP)~\cite{eriksson2021high} as the surrogate model: 
As an intuitive explanation, SAAS-GP places strong, sparsity-inducing priors on the GP hyperparameters to alleviate the difficulty in modelling high-dimensional data by assuming that despite the high nominal dimensionality, \textit{some} search dimensions contribute much more significantly to the variation of the objective function than others -- this assumption is shown to hold in related problems of \textit{NAS in computer vision} ~\cite{DBLP:conf/iclr/WanREL22} and \textit{discrete prompt search in PLMs}~\cite{zhou2023survival}, and we expect similar findings in our particular case.

For the acquisition function, we use the noisy expected hypervolume improvement (NEHVI)~\cite{daulton2021parallel} to handle the multi-objective setting: unlike the commonly used scalarisation approach that transforms the vector-valued objective function to a scalar weighted sum (which corresponds to \textit{a single point} on the PF), NEHVI is capable of automatically exploring all parts of the PF in a single run.
Lastly, we additionally use \textit{low-fidelity} approximations, a popular low-cost performance estimation strategy in NAS \cite{elsken2019neural}, to manage the search cost: at search-time, instead of fine-tuning each candidate PEFT configuration in full, we only fine-tune with a much smaller number of iterations (5\% of full) -- this is possible as we are only interested in the \emph{relative ranking} (rather than the performance itself) of the different configurations during search. Consistent with NAS literature, we also find the low-fidelity estimate to provide a reliable ranking, with the best-performing configurations in low fidelity also performing the best under fine-tuning with the full number of iterations.
As we will show in §\ref{sec:results}, using the low-fidelity search pipeline, in combination with the strong transferability of the discovered configurations, \method only incurs an additional \emph{one-off}, \textbf{1.9\%} of the total GLUE fine-tuning cost, but delivers significant performance gains.

\section{Related Work}
\label{sec:rw}
\noindent \textbf{PEFT Methods in NLP.}
Standard PEFT methods can be divided into two main groups \cite{Pfeiffer:2023survey}. \textbf{1)} Some methods fine-tune a small portion of pretrained parameters~\cite{zhao-etal-2020-masking, guo-etal-2021-parameter}. For instance, \citet{ben-zaken-etal-2022-bitfit} propose to fine-tune the PLM's bias terms, while \citet{DBLP:conf/nips/SungNR21} and \citet{ansell-etal-2022-composable} fine-tune sparse subnetworks withing the original PLM for a particular task. \textbf{2)} Other methods fine-tune an additional set of parameters~\cite{DBLP:journals/corr/abs-2205-05638}. Since there is no interference with the pretrained parameters, this class of PEFT modules, besides offering strong task performance, is arguably more modular; we thus focus on this class of PEFT methods in this work. The original \textit{adapter modules}~\cite{DBLP:conf/icml/HoulsbyGJMLGAG19, pfeiffer-etal-2020-mad} have a bottleneck \textit{serial} architecture which can be inserted into every Transformer layer, see Figure~\ref{fig:architecture and search space}. LoRA~\cite{DBLP:conf/iclr/HuSWALWWC22} assumes the low-rank intrinsic dimensionality of the target task and performs low-rank updates~\cite{DBLP:conf/nips/MahabadiHR21}. \citet{li-liang-2021-prefix} propose the Prefix-Tuning method that appends a learnable vector to the attention heads at each Transformer layer. Similarly, prompt-tuning~\cite{lester-etal-2021-power} only appends this vector to the input embedding. UniPELT~\cite{mao-etal-2022-unipelt} integrates multiple PEFT modules with a dynamic gating mechanism. \citet{DBLP:conf/iclr/HeZMBN22} provide a unified formulation of existing PEFT modules and propose a \textit{parallel} adapter module, along with a combined `Mix-and-Match Adapter (MAM)' architecture that blends parallel adapters and prefix-tuning. \citet{https://doi.org/10.48550/arxiv.2205.12410} propose the mixture-of-adaptations (AdaMix) architecture with weight averaging for a mixture of adapters. 

\vspace{1.2mm}
\noindent \textbf{Optimising Parameter Efficiency in PEFT.} Recent work further aims to optimise the parameter efficiency of existing PEFT modules while maintaining task performance. The standard approach is to insert (typically serial) adapters into all Transformer layers, which still requires a sizeable parameter budget. \citet{ruckle-etal-2021-adapterdrop} address this question by randomly dropping adapters from lower-level layers, displaying only a small decrease in task performance. Adaptable Adapters (AA)~\cite{moosavi-etal-2022-adaptable} generalise this idea by learning gates that switch on or off adapters in particular Transformer layers. Neural Architecture Search (NAS) methods aim to automate the design of neural net architectures themselves, and NAS has seen great advances recently, with performance often surpassing human expert-designed architectures in various tasks~\cite{zoph2016neural, ren2021comprehensive, elsken2019neural}. Concerning NLP tasks and PEFT, \citet{DBLP:journals/corr/abs-2206-07382} propose S\textsuperscript{3}PET, which adapts Differentiable Architecture Search (DARTS)~\cite{liu2018darts} to learn the positions for inserting the PEFT modules. This work is closest in spirit to ours and is empirically compared to in \S\ref{sec:experiments}. Conceptually, however, as discussed in detail in \S\ref{sec:autopeft}, we argue that our method offers a spectrum of advantages over S\textsuperscript{3}PET and other related PEFT work, including but not limited to the ability to automatically discover a family of PEFT configurations across parameter budgets in a single run, better parallelisability and memory efficiency.
Other concurrent work \citep{DBLP:journals/corr/abs-2210-07558, zhang2023adaptive} also approaches the same problem by dynamic budget allocation mechanisms on a single PEFT module within a limited search space. Nonetheless, this field still lacks a compact solution for automatically configuring a complex space of PEFT modules \citep{chen2023parameterefficient}.

\section{Experimental Setup}

\begin{table*}[!t]
\centering
\def\arraystretch{0.9}
\resizebox{\textwidth}{!}{%
    \begin{tabular}{@{}lcccccccccc@{}}
    \toprule
    \textbf{Method}           & \textbf{\#Param.} & \textbf{RTE} & \textbf{MRPC} & \textbf{STS-B} & \textbf{CoLA} & \textbf{SST-2} & \textbf{QNLI} & \textbf{QQP} & \textbf{MNLI} & \textbf{Avg.} \\\midrule
    
   FFT  & 100\%        &\color{lgrey}\textbf{71.12}\textsubscript{1.46}     &\color{lgrey}{85.74}\textsubscript{1.75}      &\color{lgrey}89.00\textsubscript{0.45}       &\color{lgrey}59.32\textsubscript{0.62}      &\textbf{92.57}\textsubscript{0.24}       &\textbf{\color{mgrey}91.50}\textsubscript{0.08}      &\textbf{91.52}\textsubscript{0.04}     &\textbf{84.43}\textsubscript{0.22}      &\color{lgrey}\textbf{83.15}      \\
    
    Prefix    & 0.17\%           &\color{lgrey}{70.54}\textsubscript{0.49}     &\color{lgrey}{85.93}\textsubscript{0.89}      &\color{lgrey}88.76\textsubscript{0.15}       &\color{lgrey}58.88\textsubscript{1.15}      &\color{lgrey}91.93\textsubscript{0.45}       &\color{lgrey}90.76\textsubscript{0.14}      & \color{lgrey}89.12\textsubscript{0.07}    & \color{lgrey}82.78\textsubscript{0.16}     & \color{lgrey}82.33     \\
    
    LoRA             &   0.27\%   & \color{lgrey}{65.85}\textsubscript{1.49}    & \color{lgrey}{84.46}\textsubscript{1.04}     &\color{lgrey}88.73\textsubscript{0.08}       & \color{lgrey}57.58\textsubscript{0.78}     &\color{lgrey}92.06\textsubscript{0.38}       &\color{lgrey}90.62\textsubscript{0.22}      & \color{lgrey}89.41 \textsubscript{0.04}    &\color{lgrey}83.00\textsubscript{0.07}      & \color{lgrey}81.46     \\
    
    Serial   &    0.81\%        &\color{lgrey}{68.01}\textsubscript{1.34}&\color{lgrey}{84.75}\textsubscript{0.45}      &\color{lgrey}88.61\textsubscript{0.11}      &\color{lgrey}59.73\textsubscript{0.62}  &\color{lgrey}91.93\textsubscript{0.33}      &\color{lgrey}91.06\textsubscript{0.12}       &\color{lgrey}90.52\textsubscript{0.05}      & \color{lgrey}84.18\textsubscript{0.22}    &\color{lgrey}82.35            \\
    
    AdaMix   &    0.81\%        &\color{lgrey}{70.11}\textsubscript{0.62}&\color{lgrey}\textbf{86.86}\textsubscript{1.12}      &\color{lgrey}\textbf{89.12}\textsubscript{0.11}      &\color{lgrey}59.11\textsubscript{1.00}  &\color{lgrey}92.06\textsubscript{0.22}      &\textbf{91.52}\textsubscript{0.15}       &\color{lgrey}90.22\textsubscript{0.04}      & \color{lgrey}\textbf{84.25}\textsubscript{0.14}    &\color{lgrey}82.91            \\
    
    UniPELT          &  1.25\%          &\color{lgrey}{67.07}\textsubscript{1.82}&\color{lgrey}{84.22}\textsubscript{0.78}      &\color{lgrey}88.84\textsubscript{0.11}       &\color{lgrey}\textbf{60.13}\textsubscript{0.46}      &\color{mgrey}\textbf{92.52}\textsubscript{0.24}       &\color{lgrey}91.09\textsubscript{0.13}     &\color{lgrey}90.69\textsubscript{0.11}     &\color{mgrey}\textbf{84.28}\textsubscript{0.18}      &\color{lgrey}82.35      \\
    
    Parallel &   6.46\%         &\color{lgrey}{68.52}\textsubscript{3.44}     &\color{lgrey}{86.52}\textsubscript{0.96}      &\color{lgrey}88.90\textsubscript{0.28}       & \color{lgrey}58.72\textsubscript{1.69}     & \color{lgrey}92.13\textsubscript{0.35}      &\color{lgrey}90.83\textsubscript{0.22}      &  \color{lgrey}\textbf{90.74}\textsubscript{0.08}   &   \color{lgrey}73.93\textsubscript{19.24}   &  \color{lgrey}81.29    \\
    
    MAM      &   6.97\%         &\color{lgrey}{69.10}\textsubscript{1.76}     & \textbf{\color{mgrey}{87.16}}\textsubscript{0.74}     &\color{lgrey}\textbf{89.01}\textsubscript{0.48}       &\color{lgrey}47.87\textsubscript{23.97}      &\color{lgrey}83.94\textsubscript{16.52}       &\color{lgrey}90.85\textsubscript{0.22}      &\color{mgrey}\textbf{90.76}\textsubscript{0.05}     &\color{lgrey}83.31\textsubscript{0.17}      &\color{lgrey}80.25     \\ \midrule
    
    $\method^{\textit{RTE}}$     &  0.76\%          &\color{mgrey}\textbf{72.20}\textsubscript{0.72}     &\color{mgrey}\textbf{87.16}\textsubscript{0.83}      &\color{lgrey}88.77\textsubscript{0.07}       &\color{mgrey}\textbf{60.30}\textsubscript{1.24}      &\color{lgrey}\textbf{92.22}\textsubscript{0.30}       &\color{lgrey}90.90\textsubscript{0.10}      & \color{lgrey}90.37\textsubscript{0.06}    & \color{lgrey}83.46\textsubscript{0.21}     & \color{mgrey}\textbf{83.17}     \\ \midrule
    
    $\method^{\textit{task}}_{\textit{Avg.}}$     & $\overline{1.40}$\%          &\textbf{72.35}\textsubscript{0.94}     &\textbf{87.45}\textsubscript{0.87}      &\textbf{89.17}\textsubscript{0.24}      & \textbf{60.92}\textsubscript{1.47}    & \color{lgrey}\textbf{92.22}\textsubscript{0.30}    & \color{lgrey}\textbf{91.12}\textsubscript{0.13} & \color{lgrey}90.64\textsubscript{0.05}& \color{lgrey}84.01\textsubscript{0.10}&   \textbf{83.49}   \\ \bottomrule
\end{tabular}}
\caption{Results on the GLUE benchmark with BERT\textsubscript{base} (tasks are ranked in ascending order of training resources required from left to right). For $\method^{\textit{RTE}}$, we search on RTE with a low-fidelity proxy, training for 1 epoch per iteration, \emph{only at a search cost of 1.9\% (in terms of additional fine-tuning steps required) over the full GLUE experiment}. We report the $\overline{\text{average}}$ fine-tuned parameters of \textit{per-task} \method, where we conduct additional \textit{per-task} searches on RTE, MRPC, STS-B, and CoLA, and take best-found configurations for the remaining tasks. We report Spearman's Correlation for STS-B, Matthew's Correlation for CoLA, and accuracy for all other tasks (matched accuracy for MNLI). The percentage of parameters is the ratio of the number of additional parameters to the pretrained parameters. We reproduce all baselines and report the mean and standard deviation of all results for 5 random seeds. The \textbf{best}, \textbf{\color{mgrey}{second-best}}, and \textbf{\color{lgrey}third-best} results are marked in bold fonts and ranked by colour.}
\label{tab:keyresults}
\end{table*}
\begin{table}[!t]
\centering
\small
\resizebox{\linewidth}{!}{%
\begin{tabular}{@{}lcclc@{}}
\toprule
\textbf{Task}                & \textbf{\%Param.}                   & \multicolumn{1}{c}{\textbf{Active PEFT }}          & \textbf{Submodule}      & \multicolumn{1}{l}{\textbf{Value}} \\ 
{} & {} & \multicolumn{1}{c}{\textbf{Layers} $l_i$} \\
\midrule
    \multirow{3}{*}{RTE} & \multirow{3}{*}{0.76\%} & \multirow{3}{*}{\shortstack{3, 4,\\ 8, 9, 10}} & $D_{\text{SA}}$ (Serial)  & 12                            \\
 &  &    &  $D_{\text{PA}}$ (Parallel) & 96 \\
 &  &    & $L_{\text{PT}}$ (Prefix)    & 1  \\\bottomrule
\end{tabular}}
\vspace{-2mm}
  \caption{Specification of the discovered configuration reported in Table~\ref{tab:keyresults} ($\method^{\textit{RTE}}$) using BERT\textsubscript{base}.}
 \label{tab:autopeftfoundconfigs}
 \vspace{-2mm}
\end{table}
\label{sec:experiments}
\noindent \textbf{Evaluation Data.}
We follow prior PEFT research and base our evaluation on the standard and established GLUE and {SuperGLUE benchmarks}. For GLUE, we include 4 types of text classification tasks, including linguistic acceptability: CoLA; similarity and paraphrase: STS-B, MRPC, QQP; sentiment analysis: SST-2; natural language inference: RTE, QNLI, MNLI. We exclude WNLI following previous work~\cite{DBLP:conf/icml/HoulsbyGJMLGAG19, mao-etal-2022-unipelt}. We also include CB, COPA, WiC, and BoolQ from SuperGLUE to further validate the transferability of \method-found configuration across different tasks and datasets.

\vspace{1.3mm}
\noindent \textbf{Baselines.}
We compare the performance of the \method-found configurations to the standard full model FT and each individual PEFT module (SA, PA, PT) from the search space used in their default setup from their respective original work. We also compare with the LoRA module to provide a comparison to low-rank decomposition methods. To compare with recent methods that also integrate multiple PEFT modules (see \S\ref{sec:rw}), we further include the UniPELT and the MAM adapter in their default settings. We reproduce AdaMix for a comparison to a mixture of homogeneous adaptations. In ablations on insertion layers, we also include the Adaptable Adapter (AA) as a baseline that proposes a differentiable gate learning method to select the insertion layer for PEFT modules (i.e. serial adapters originally). On T5 \citep{raffel2020t5} models, we also compare against S\textsuperscript{3}PET \cite{DBLP:journals/corr/abs-2206-07382}, one of the most similar works to us that use differentiable NAS for configuration search.

\vspace{1.3mm}
\noindent \textbf {Implementation Details.}
Following previous work on the GLUE benchmark, we report the best GLUE dev set performance~\cite{ben-zaken-etal-2022-bitfit} and use 20 training epochs with an early stopping scheme of 10 epochs for all \textit{per-task} experiments. We use AdapterHub~\cite{pfeiffer-etal-2020-adapterhub} as the codebase and conduct extensive experiments with the uncased BERT\textsubscript{base}~\cite{devlin-etal-2019-bert} as the main backbone model. We report main experiments with the mean and standard deviation over 5 different random seeds. {Following \citet{pfeiffer-etal-2020-mad}, we use a recommended learning rate of $10^{-4}$ for all PEFT experiments. We use the learning rate of $2 
\times 10^{-5}$ for full model FT according to~\citet{mao-etal-2022-unipelt}. We use batch sizes 32 and 16 for all BERT and RoBERTa experiments, respectively. The optimiser settings for each PEFT module follow the default settings in AdapterHub~\cite{pfeiffer-etal-2020-adapterhub}. We implement the BO search algorithm in BoTorch \cite{balandat2020botorch} and use the recommended settings from \citet{eriksson2021high} for the surrogate. For acquisition function optimisation, we use a local search method similar to previous literature with a similar setup \cite{wan2021think, eriksson2021latency}: at each search iteration (after the initial randomly sampled points), we collect the \textit{Pareto-optimal} architectures up to this point. From this collection of Pareto-optimal architectures, we perform a local search by evaluating the acquisition function values of their neighbours and move the current point to a neighbour with a higher acquisition function value, and this process is repeated until convergence. Due to the relatively noisy nature of the problem, we use 100 random initialisation points for all experiments, followed by 100 BO iterations. We further show results using RoBERTa\textsubscript{large}~\cite{DBLP:journals/corr/abs-1907-11692} in Table~\ref{tab:keyresults_roberta}, which shows findings that are consistent with the BERT\textsubscript{base}. In experiments with RoBERTa\textsubscript{large} as the underlying PLM, we report the RTE results with a learning rate of  $2 
\times 10^{-5}$ for $\method^{\textit{MRPC}}$ and $\method^{\textit{CoLA}}$; $10^{-4}$ for $\method^{\textit{RTE}}$. We use batch size 16 and a learning rate of $3 
\times 10^{-4}$ for T5\textsubscript{base} experiments by \method with the SAPA space; $10^{-5}$ for STS-B. We reproduce S\textsuperscript{3}PET results with batch size 8 in the same experimental setup as \method.

\begin{table}[!t]
\centering
\def\arraystretch{0.9}
\resizebox{\linewidth}{!}{%
\begin{tabular}{@{}lccccc@{}}
\toprule
\textbf{Method}            & \textbf{CB} & \textbf{COPA} & \textbf{WiC} & \textbf{BoolQ}  & \textbf{Avg.} \\ \midrule
FFT           &\textbf{71.43}\textsubscript{1.13}         &\color{lgrey}51.80\textsubscript{3.76}      & \color{mgrey}\textbf{68.62}\textsubscript{1.93}     & \textbf{72.17}\textsubscript{0.86} & \color{mgrey}\textbf{66.01}        \\
LoRA          &\color{lgrey}67.14\textsubscript{2.42}         &\color{mgrey}\textbf{55.80}\textsubscript{1.47}      & \color{lgrey}\textbf{68.56}\textsubscript{1.11}     & \color{lgrey}\textbf{69.09}\textsubscript{0.42} & \color{lgrey}\textbf{65.15}        \\

Serial           &\color{lgrey}\textbf{67.86}\textsubscript{1.13}         &\color{lgrey}\textbf{54.20}\textsubscript{7.68}      & \color{lgrey}67.34\textsubscript{0.61}     & \color{lgrey}70.00\textsubscript{0.85} & \color{lgrey}64.86        \\
Ours$^{\textit{RTE}}$            &\color{mgrey}\textbf{71.07}\textsubscript{2.86}     &  \textbf{56.40}\textsubscript{6.83}    &  \textbf{68.87}\textsubscript{1.06}     &  \color{mgrey}\textbf{70.86}\textsubscript{0.89}    &   \textbf{66.80}      \\ 
\bottomrule
\end{tabular}}
\caption{{Results on SuperGLUE tasks with \method-discovered configurations \emph{searched on RTE} with BERT\textsubscript{base} as the underlying PLM. We split 10\% of the training set as the new validation set and report the $\method^{\textit{RTE}}$-found configuration transfer results on the evaluation set over five random seeds.}}
\label{tab:keyresults_superglue}
\end{table}
\begin{figure*}[!t]
    \centering
    \includegraphics[width=0.999\linewidth]{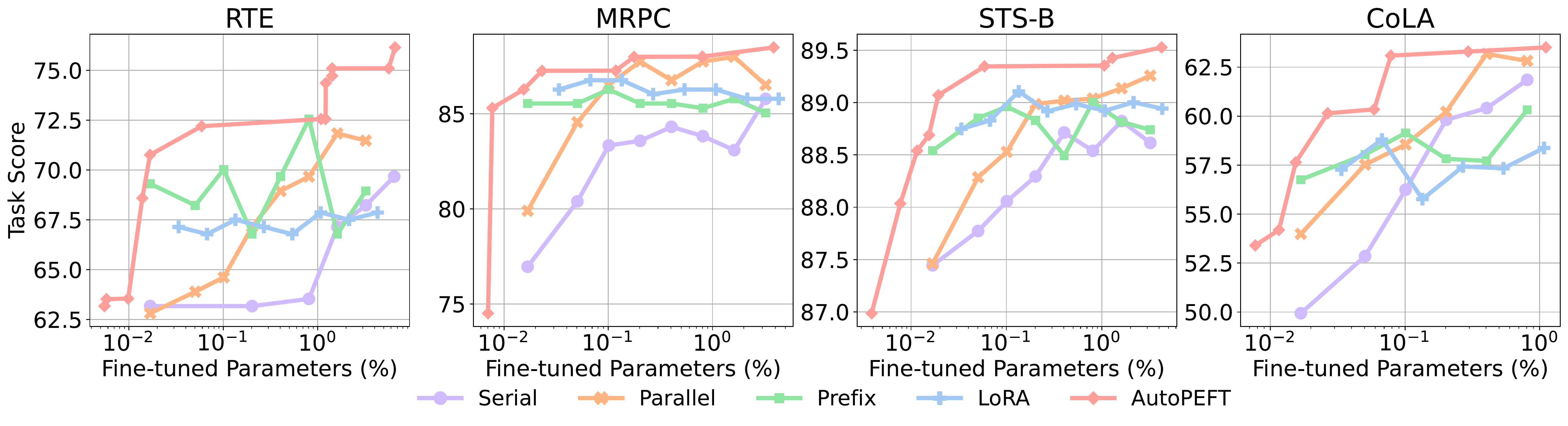}
    \vspace{-7mm}
    \caption{Pareto Fronts of \method on four tasks compared to baselines on BERT\textsubscript{base}, over varying parameter budgets. We report the single-seed task score but otherwise follow the settings in Table~\ref{tab:keyresults}.}
    \label{fig:searchwithbaselines}
\end{figure*}

\section{Results and Discussion}
\label{sec:results}

\begin{table*}[!t]
\centering
\def\arraystretch{0.9}
\resizebox{0.9\linewidth}{!}{%
\begin{tabular}{@{}lcccccccccccc@{}}
\toprule
\textbf{Method} & \textbf{\#Param.}           & \textbf{RTE} & \textbf{MRPC} & \textbf{STS-B} & \textbf{CoLA} & \textbf{SST-2} & \textbf{QNLI} & \textbf{QQP}& \textbf{MNLI}& \textbf{Avg.} \\ \midrule

LoRA  &    0.40\%         &\color{mgrey}\textbf{80.1}         &\textbf{89.5}      & \color{lgrey}\textbf{89.2}     & \color{lgrey}\textbf{59.9} & \color{lgrey}\textbf{94.4}    &   \textbf{93.6}      &       \textbf{91.0} &\color{mgrey}\textbf{86.5}& \color{mgrey}\textbf{85.5}  \\

Serial    &    0.79\%       &\color{lgrey}78.0         &\color{lgrey}88.2      & \color{lgrey} 89.1    & \color{mgrey}\textbf{60.6} & \textbf{94.6}    &   \color{mgrey}93.1     &       \color{lgrey}\textbf{90.7} &\color{lgrey}86.4&\color{lgrey}85.1   \\ \midrule

S\textsuperscript{3}PET$^{\textit{RTE}}$  & 0.30\%        &\color{lgrey}\textbf{79.8}     &\color{mgrey}\textbf{89.0}      &\textbf{90.2}       & \color{lgrey}58.6  &\color{mgrey}\textbf{94.2}       &\color{mgrey}\textbf{93.3}           & \color{lgrey}90.6   &\color{mgrey}\textbf{86.5}& \color{lgrey}\textbf{85.3} \\

$\method^{\textit{RTE}}$    &  0.33\%           &\textbf{82.7}     &  \color{mgrey}\textbf{89.0}    &  \color{mgrey}\textbf{89.6}     &   \textbf{61.7}   &  \textbf{94.6}     &  \color{mgrey}\textbf{93.3}         & \color{mgrey}\textbf{90.8}  &\textbf{86.7}& \textbf{86.1}\\ 
\bottomrule
\end{tabular}
}
\caption{Experimental results on GLUE with T5\textsubscript{base}. We report comparisons of in-task search performance and transfer performance between the architectures found by \method and the state-of-the-art baseline S\textsuperscript{3}PET in a constrained parameter budget. Consistent with Table \ref{tab:keyresults},  we report \method and S\textsuperscript{3}PET results searched on RTE in full-resource settings that are then transferred to all other included GLUE tasks.}
\label{tab:keyresults_t5}
\end{table*}

\begin{table*}[!t]
\centering
\def\arraystretch{0.83}
\resizebox{0.75\linewidth}{!}{%
\begin{tabular}{@{}lcccccccccc@{}}
\toprule
\textbf{Method} & \textbf{\#Param.}           & \textbf{RTE} & \textbf{MRPC} & \textbf{STS-B} & \textbf{CoLA} & \textbf{SST-2} & \textbf{QNLI} & \textbf{Avg.} \\ \midrule
FFT$^{\dag}$  & 100\%        &\color{mgrey}\textbf{86.6}     &\color{mgrey}\textbf{90.9}      &\textbf{92.4}       & \color{lgrey}\textbf{68.0}  &\color{mgrey}\textbf{96.4}       &\color{mgrey}\textbf{94.7}           & \color{lgrey}\textbf{88.2}     \\

LoRA$^{\ddag}$  &    0.22\%         &\color{lgrey}\textbf{85.2}         &\color{lgrey}\textbf{90.2}      & \color{mgrey}\textbf{92.3}     & \color{mgrey}\textbf{68.2} & \color{lgrey}96.2    &   \textbf{94.8}      &       \color{lgrey}87.8    \\

Serial    &    0.89\%       &\color{lgrey}84.8         &\color{lgrey}\textbf{90.2}      & \color{lgrey}\textbf{92.0}     & \color{lgrey}66.8 & \color{lgrey}\textbf{96.3}    &   \color{mgrey}\textbf{94.7}      &       \color{lgrey}87.5    \\
\midrule

$\method^{\textit{RTE}}$    &  0.03\%           &\textbf{88.1}     &  \color{lgrey}89.5    &  \color{mgrey}\textbf{92.3}     &   \color{lgrey}67.0   &  \color{lgrey}96.0     &  \color{lgrey}\textbf{94.6}         & \color{lgrey}\textbf{87.9}   \\ \midrule
$\method^{\textit{task}}_{\textit{Avg.}}$  &$\overline{0.88}$\%            &\textbf{88.1}     &  \textbf{92.2}    &  \textbf{92.4}     &  \textbf{70.6}    &   \textbf{96.8}    &   \color{lgrey}\textbf{94.6} &  \textbf{89.1}   \\ 
\bottomrule
\end{tabular}
}
\caption{Experimental results on GLUE with RoBERTa\textsubscript{large}. We report the full model fine-tuning$^{\dag}$ results from~\citet{DBLP:journals/corr/abs-1907-11692} with Pearson correlation for STS-B. We include the LoRA$^{\ddag}$ module performance from~\citet{DBLP:conf/iclr/HuSWALWWC22}. We exclude QQP and MNLI tasks due to the high computation cost of RoBERTa\textsubscript{large}. Consistent with Table \ref{tab:keyresults},  we again report \method results searched on RTE in full-resource settings that are then transferred all included GLUE tasks ($\method^{\textit{RTE}}$) and per-task \method ($\method^{\textit{task}}_{\textit{Avg.}}$) but on RoBERTa\textsubscript{large}.}
\label{tab:keyresults_roberta}
\end{table*}

\noindent \textbf{Discussion of Main Results.}
{
The main results on BERT are summarised in Table~\ref{tab:keyresults} where we evaluate the \method-found configurations searched from RTE, the most low-resource and challenging task, on the full GLUE suite. We further report selected GLUE tasks on T5 in Table~\ref{tab:keyresults_t5} (where we also compare against S\textsuperscript{3}PET) and RoBERTa\textsubscript{large} in Table~\ref{tab:keyresults_roberta}. For simplicity, we report a single configuration that leads to the highest task performance in a predefined, user-specified parameter budget from the discovered Pareto-optimal set in Table \ref{tab:keyresults}, whereas the full Pareto-optimal set is evaluated in Figure \ref{fig:searchwithbaselines}}. On BERT (Table~\ref{tab:keyresults}, we find that using only 0.76\% of parameters, $\method^{\textit{RTE}}$ outperforms all the PEFT baselines (more than 2\% on  RTE). The \method-found configuration also outperforms the full-model FT baseline on the RTE task by more than 1\%. These results indicate the effectiveness of the \method framework in optimising both task performance and parameter efficiency. Transferring the RTE-based configurations to other tasks, we find that strong performance is maintained across the target tasks, with more benefits on the medium-resource tasks (MRPC, STS-B, CoLA), but the configuration remains competitive also for higher-resource tasks (e.g. QQP, MNLI). Finally, we find the strength of \method to persist in RoBERTa and T5 as a representative of the encoder-decoder model families. It is particularly noteworthy that in addition to outperforming the baseline PEFT methods without configuration search, \method also performs competitively compared to S\textsuperscript{3}PET \textit{with configuration search} under a comparable parameter count, even though S\textsuperscript{3}PET was \textit{exclusively developed and tested on the T5 search space} and that the \textit{S\textsuperscript{3}PET search space was designed with meticulous hand-tuning}, where the authors manually excluded several building blocks that did not lead to empirical gain; this provides further empirical support to the strength of a BO-based search strategy described in \S\ref{sec:bo_method}.

{Table~\ref{tab:autopeftfoundconfigs} specifies the composition of the found configuration, indicating the exact task-active layers while allocating more parameter budget to the efficient and effective PA module.} On average, the $\method^{\textit{RTE}}$ configuration shows a comparable fine-tuning performance (83.17) to FFT (83.15) by only updating 0.76\% of parameters. With strong transferability across similar tasks, \method provides distinct advantages in parameter efficiency; the search algorithm itself, coupled with the transfer, becomes more sample-efficient within limited training resources.

\begin{figure}[!t]
    \centering
    \includegraphics[width=0.8\linewidth]{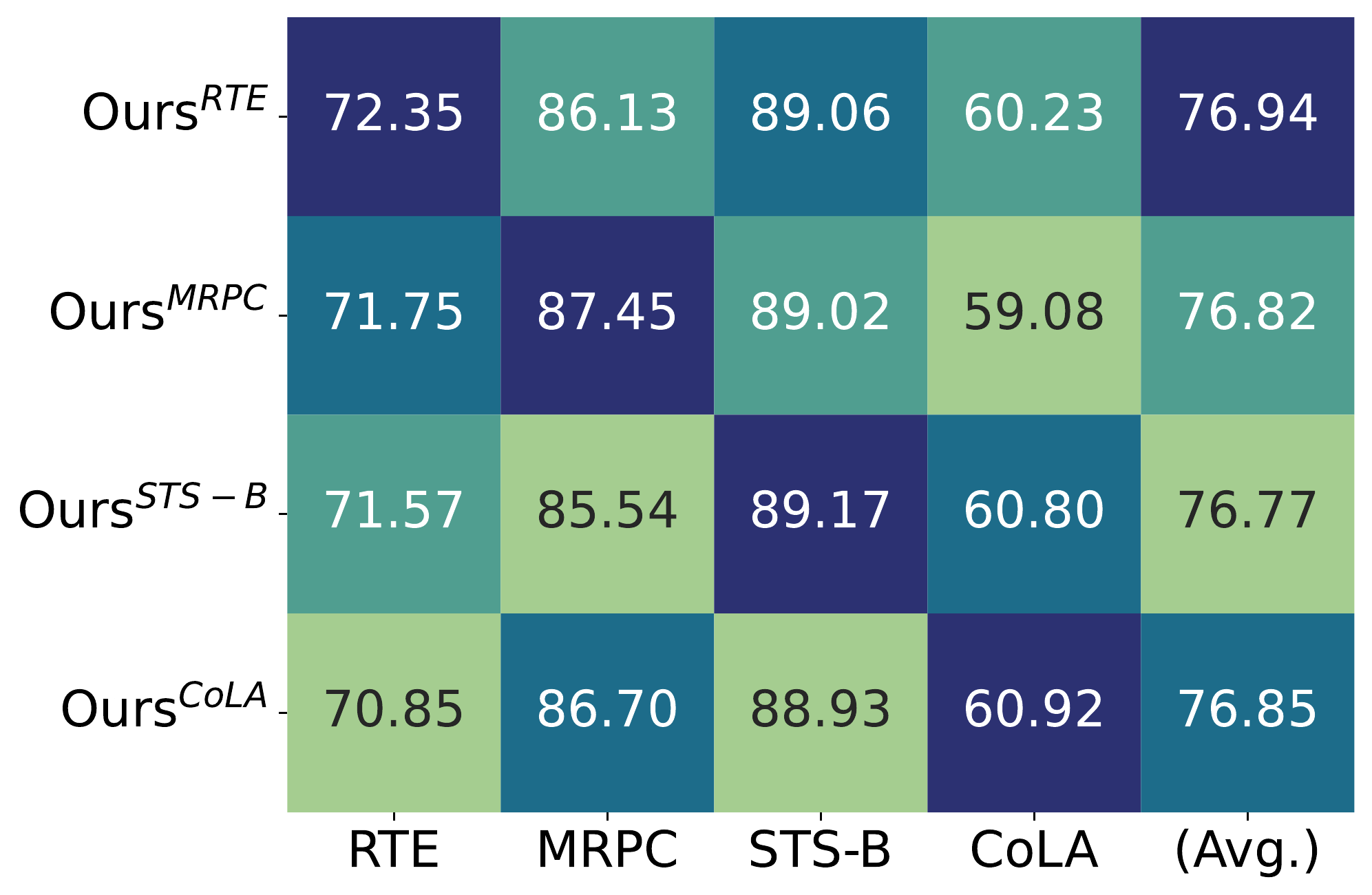}
    \vspace{-2mm}
    \caption{{Pairwise transferability study of \method-discovered configurations: each \textbf{row} (\texttt{Ours\textsuperscript{[task]}}) denotes the performances of the \method configuration searched from \texttt{[task]} (e.g. RTE) to the task itself and 3 other GLUE tasks. The results suggest that \method performance is largely robust to the choice of which task to search on.}}
    \label{fig:heatmap}
\end{figure}
\vspace{1.2mm}
\noindent \textbf{Extending \method to More Tasks.}
{We next `stress-test' the ability of \method-found configuration in a more challenging scenario, experimenting on a completely new set of dissimilar tasks. Table~\ref{tab:keyresults_superglue} reports the results of transferring $\method^{\textit{RTE}}$ from Table \ref{tab:keyresults} to four SuperGLUE tasks.
In terms of \emph{parameter efficiency}, we observe consistent patterns as in Table~\ref{tab:keyresults} before, where our \textit{plug-and-play} PEFT configuration outperforms existing PEFT baselines by a substantial margin (2\%) on average while being comparable to the costly full-model FT. \footnote{With the \method-found off-the-shelf configuration, this requires no additional search cost and enables a more efficient and effective tuning approach for new tasks.}In terms of \emph{search cost}, we recall that through the use of low-fidelity proxy and the strong transferability, $\method^{\textit{RTE}}$ in Table~\ref{tab:keyresults} only requires an additional, one-off 1.9\% in terms of training time (or equivalently the number of fine-tuning steps) of that of single-seed training of the GLUE training sets. Furthermore, Figure~\ref{fig:heatmap} demonstrates the robustness of our framework to the choice of the source task to search on. Therefore, our framework is task-agnostic with a cheap one-time cost but yields `permanent' improvement towards all efficiency metrics for PEFT: space, time, and memory.}

\vspace{1.2mm}
\noindent \textbf{\textit{Per-Task} Search.}
We further conduct full-resource per-task \method searches. While naturally more expensive, we argue this setup is useful if, for example, one is interested in finding absolutely the best configurations \emph{for that particular task} and where search cost is less of a concern. Due to computational constraints, we search per-task on RTE, MPRC, STS-B, and CoLA, then port the small set of best configurations to the remaining higher-resource tasks (SST-2, QNLI, QQP, MNLI). We observe consistent gain in all tasks we search on over the best-performing PEFT baselines, e.g. MRPC (87.16\% (\textit{best baseline}) to 87.45\% (\textit{ours})) and CoLA (60.13\% to 60.92\%), and also the transferred configuration $\method^{\textit{RTE}}$ in Table \ref{tab:keyresults}. One interpretation is that while configurations are highly transferable, the optimal configurations may nonetheless differ slightly across tasks such that while transferred \method configurations (e.g. the one reported in Table \ref{tab:keyresults}) perform \emph{well}, searching per-task performs the \emph{best}. Crucially, we also find per-task \method in this setup to even \textit{outperform FFT, despite only using 1.4\% of all parameters}, except for the high-resources task where we mostly perform on par; this is consistent with our observations that similar to the baselines, due to the richness of training resources, the performance may be mostly saturated and PEFT methods often achieve on-par performance to FFT at most.

\begin{figure*}[!t]
    \centering
    \includegraphics[width=0.97\linewidth]{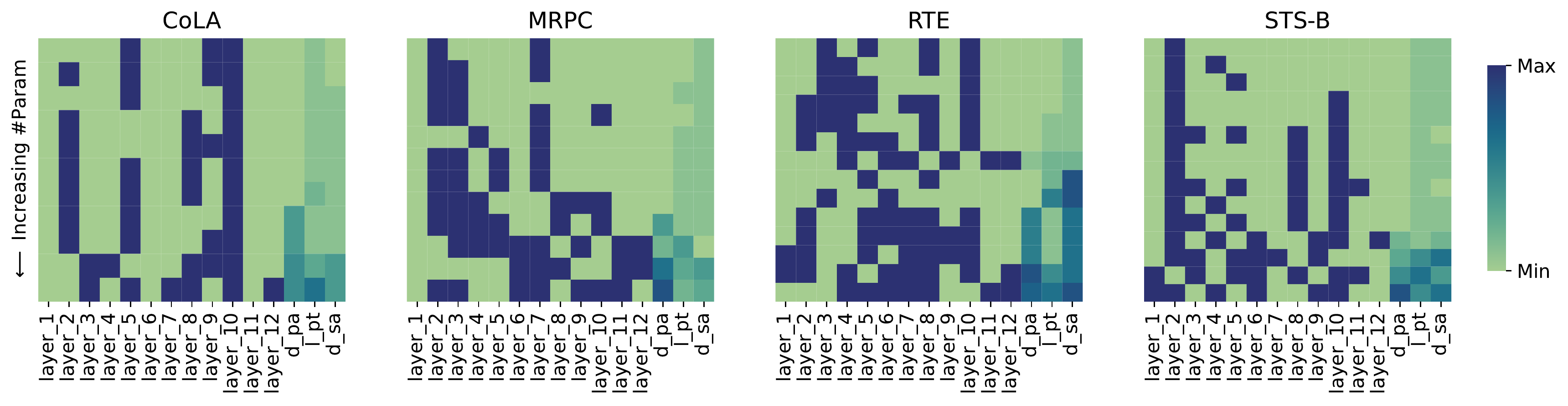}
    \vspace{-2mm}
    \caption{Visualisation of the BO discovered Pareto-optimal sets of configurations $A^*$ in different tasks (i.e., the configurations on the PFs in Figure~\ref{fig:searchwithbaselines}) in ascending order of parameter budget. \texttt{layer\_i} denotes the binary choice of whether the PEFT module is active in the $i$-th layer of the PLM. The final 3 columns denote $D_{\mathrm{SA}}, D_{\mathrm{PA}}$ and $L_{\mathrm{PT}}$ respectively, and feature a range of possible values from 0 to 768.}
    \label{fig:autopeft_pareto_configs}
\end{figure*}
\begin{figure}
    \centering
    \includegraphics[width=0.75\linewidth]{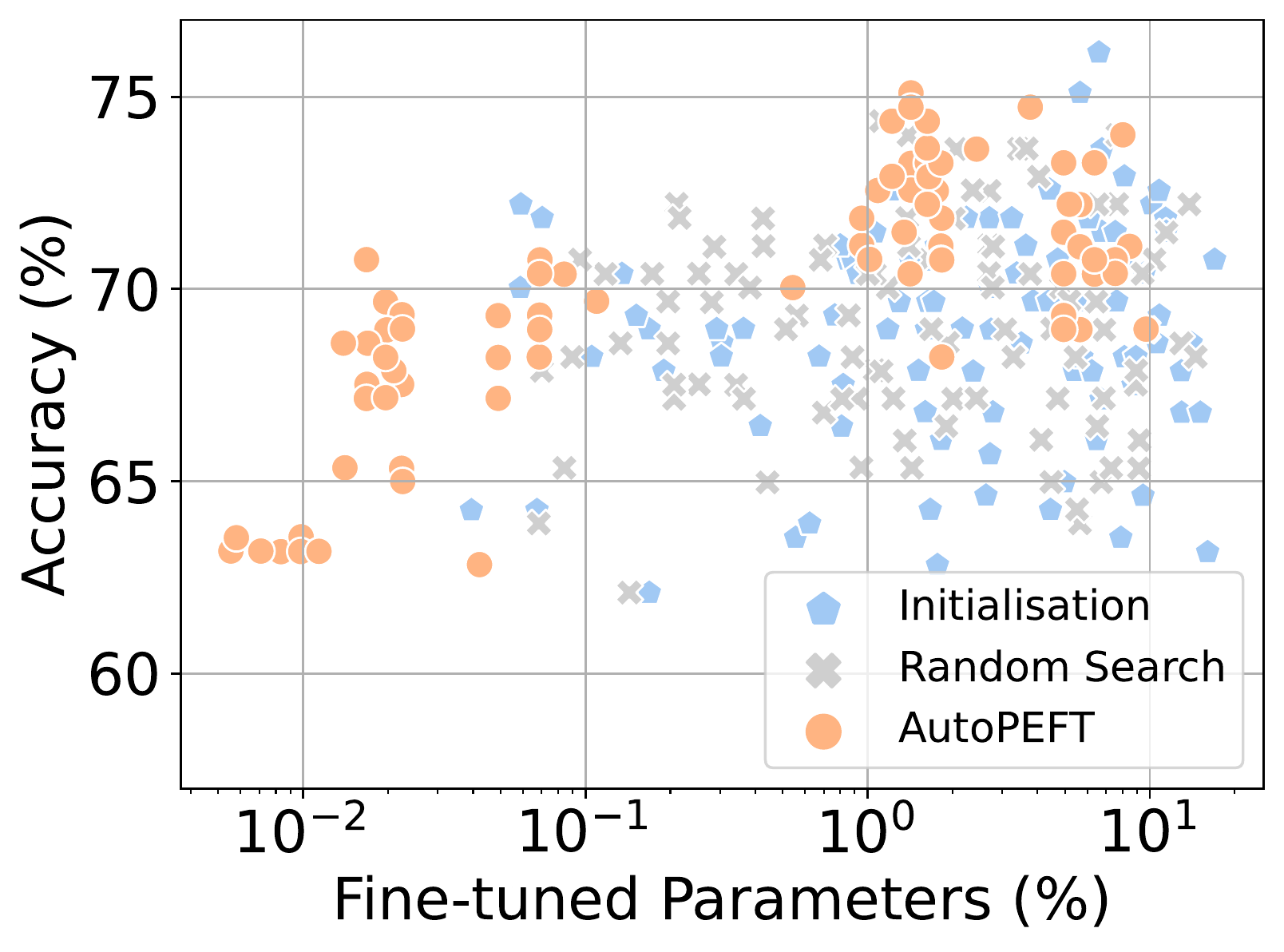}
    \vspace{-2mm}    \caption{The distribution of the discovered configurations via BO (\textcolor{orange}{orange}), described in \S\ref{sec:bo_method} and random search (\textcolor{gray}{grey}) using the same total number of evaluations (200). Both searches use the same 100 random initialising points (\textcolor{blue}{blue}) on RTE. Note that BO-generated configurations typically have much better parameter efficiency for configurations with similar accuracy.}
    \label{fig:randomsearch}
\end{figure}

\vspace{1.2mm}
\noindent \textbf{Analysing the `Behaviour' of BO and the Discovered Configurations.} Figure~\ref{fig:randomsearch} shows the distribution of \method-found configurations when we conduct its search experiment on RTE. Recalling that the search strategy (\S\ref{sec:bo_method}) starts with random initialisation, we compare the behaviours of the random explorations and the BO-suggested configurations:  whereas the random search baseline is purely exploratory and discovers less parameter-efficient configurations, BO succeeds in discovering configurations towards the regions with improved parameter efficiency. The superiority of BO over the random search baseline is further demonstrated quantitatively by Figure~\ref{fig:randomsearchvolume} where we compare the evolution of the \textit{hypervolume}, which measures the size of the space enclosed by the Pareto front over a reference point (set to the nadir point of the optimisation trajectory) \cite{zitzler1998multiobjective}, discovered by BO and random search as a function of the number of configurations evaluated; it is clear that as optimisation proceeds, BO finds a better Pareto set with a better trade-off between performance and cost in the end. BO eventually discovers a rich family of PEFT configurations across a wide range of parameters, whereas previous approaches typically fail to explore the entire PF. This is a critical strength motivating our BO search strategy. \begin{figure}
    \centering
    \includegraphics[width=0.78\linewidth]{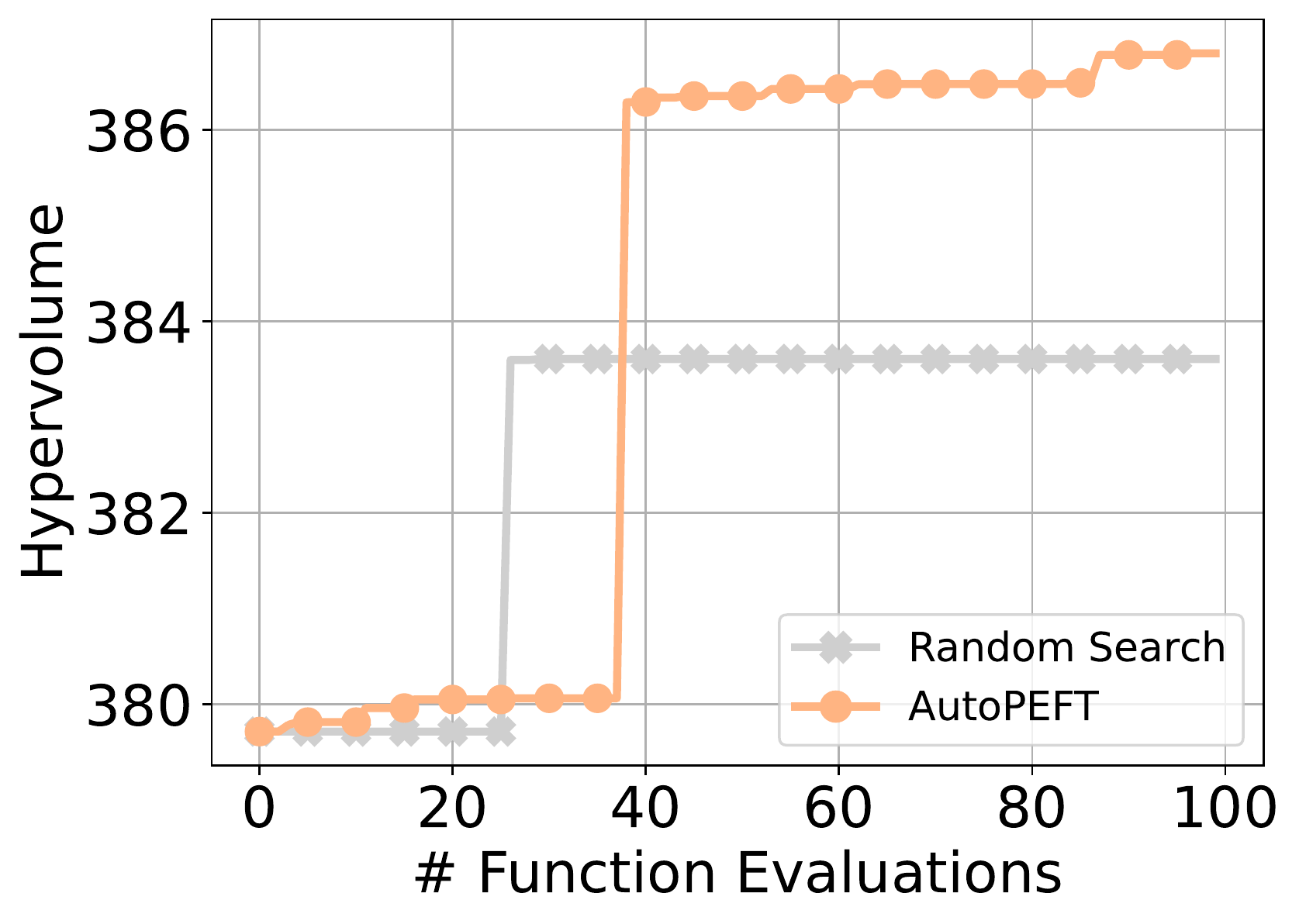}
    \vspace{-2mm}    \caption{The hypervolumes of the Pareto-optimal configurations discovered by BO (\textcolor{orange}{orange}) and random search (\textcolor{gray}{grey}) as a function of the number of configurations evaluated.}
    \label{fig:randomsearchvolume}
\end{figure}Figure~\ref{fig:autopeft_pareto_configs}, on the other hand, visualises the discovered sets in different tasks: we observe that \textit{within} the Pareto-optimal configuration set of the same task, some layers are consistently enabled (e.g., Layer 2 in CoLA) whereas some are consistently disabled (e.g., Layer 1 across all tasks) even under very different cost budgets; this suggests PEFT modules in different layers are not equally important, and by \textit{selectively} enabling them, \method is capable of making better use of the parameter budgets by allocating them to the more beneficial Transformer layers only. We observe the unanimity of preference or disinclination towards certain layers extends even \textit{across} tasks that are unlikely to stem from randomness only: for example, we found Layers 2 and 10 are enabled in 71.2\% and 69.2\% in all Pareto-optimal configurations over all tasks, whereas Layers 1 and 12 are enabled in only 7.7\% and 13.4\% of the time, respectively. We also observe that across all tasks, a common trend is that sequential and prefix adapters are universally preferred in low-budget ranges, and parallel adapters are only enabled when we have a more lenient budget allowance; these commonalities in high-performing configurations may, to some extent, account for the strong transferability of the discovered configurations, as shown in Figure~\ref{fig:heatmap}.

\begin{figure}[!t]
    \centering
    \includegraphics[width=0.8\linewidth]{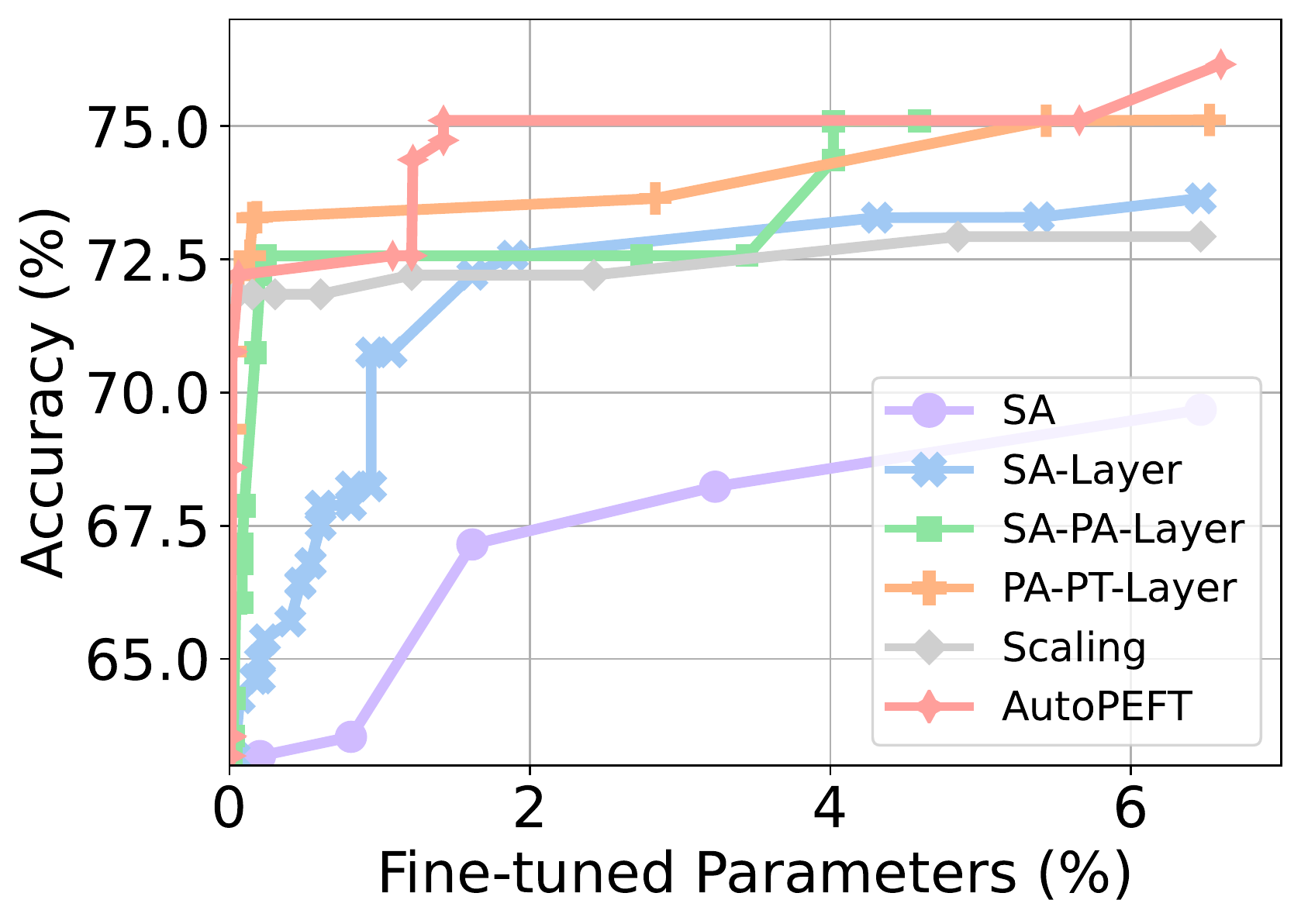}
    \vspace{-3mm}
    \caption{The performance of \method with ablation of search space on RTE on BERT\textsubscript{base}. The SA results refer to the Pfeiffer adapter~\cite{pfeiffer-etal-2020-mad} with an enumeration of its bottleneck size. The \texttt{Scaling} results refer to the PF where smaller configurations are obtained by simply scaling the largest configuration in $\mathcal{A}$ over all search dimensions. We report the PF of \method-found configurations, where \texttt{SA-PA-PT-Layer} forms the search space of \method. }
\label{fig:search_space_ablation}
\vspace{-1mm}
\end{figure}
\vspace{1.2mm}
\noindent \textbf{Ablation of the Configuration Space.} To provide a finer-grained analysis of factors that bring positive impact to \method, we ablate the \method search space from the full configuration space: 1) to the basic enumeration of the bottleneck size $D_{\text{SA}}$ of the SA only (the \texttt{SA} space); 2) a na\"ive baseline where instead of searching for each search dimension independently, we vary a single, common coefficient that generates a family of configurations of different sizes by scaling from the largest PEFT configuration in our search space (SA-PA-PT) over $D_{\text{SA}}, D_{\text{PA}}$ and $L_{\text{PT}}$. We then include the Transformer layer and the SA size into the search space (the \texttt{SA-Layer} space) to validate the usefulness of layer selection as one configuration dimension. We can then also expand the search space by adding another module (e.g. PA yields the \texttt{SA-PA-Layer} space). Figure~\ref{fig:search_space_ablation} plots the performance over the ablated configuration spaces and different parameter budgets. Several key findings emerge. First, combining multiple single PEFT modules has a positive impact on \method in general (c.f. full \method vs. \texttt{SA-PA-Layer} vs \texttt{SA-Layer}). Second, simply scaling all search dimensions by a common scaling factor is sub-optimal. This is likely because not all parameters are equally important, necessitating a configuration search. Relying on layer selection also brings benefits (c.f. \texttt{SA} vs. \texttt{SA-Layer}). The comparison indicates that \textit{leaving out Transformer layers while increasing the capacity of the PEFT module} is a straightforward method to improve the parameter efficiency and task performance of the PEFT module within a fixed parameter budget. The ablation results also demonstrate that \method is search space-agnostic, capable of effectively operating over configuration spaces of different granularity. %

\vspace{1.3mm}
\noindent \textbf{Layer Selection.}
The ability to disable some PEFT layers altogether is a key novelty of the \method search space, and to further compare different layer selection approaches, we conduct a controlled experiment with the SA module on BERT\textsubscript{large} (24 Transformer layers) under a predefined parameter budget. In Table~\ref{tab:layerselection}, we compare against AdapterDrop, which simply drops the adapters for the first 11 layers while doubling their bottleneck sizes, and, within the same architecture, we also include the Adaptable Adapter with selected layers from switch learning (3 and 10 layers from the first 12 and the other 12 layers, respectively). We show that \method outperforms existing layer selection baselines activating fewer PEFT layers, leading to better parameter efficiency (12.5\% fewer parameters in relative terms) yet achieving better performance. It indicates that selecting the best insertion layer is non-trivial, and \method can efficiently learn the correlation between layers. %

\begin{table}
\centering
\def\arraystretch{0.85}
\small
\resizebox{0.95\linewidth}{!}{%
\begin{tabularx}{\linewidth}{@{}lccX}
\toprule
\textbf{Method}  & \textbf{\#Layers} & \textbf{Size $D_{\text{SA}}$} & \textbf{RTE} \\ \midrule
Serial         & 24                & 64         & 72.56\textsubscript{0.76}         \\
Adaptable Adapter & 13                  & 128         &   73.36\textsubscript{0.80}           \\
AdapterDrop      & 13                  & 128         &   73.50\textsubscript{1.40}           \\
$\method^{\textit{SA}}_{\textit{Layer}}$             &      \textbf{10}             & 128         & \textbf{73.86}\textsubscript{0.94}        \\ \bottomrule
\end{tabularx}}
   \caption{Comparing \method to layer selection baselines with the same parameter budget on BERT\textsubscript{large}. We report the Pfeiffer adapter for all 24 layers (\texttt{Serial}), specialised \texttt{AdapterDrop}~\cite{ruckle-etal-2021-adapterdrop} that inserts SA for the last 13 layers, and AA\textsuperscript{\textit{uni}} ~\cite{moosavi-etal-2022-adaptable} without its rational activation function with 13 selected layers (\texttt{Adaptable Adapter}). We run our \method under the comparable search space of 24 layers and approximately match the size of \texttt{Serial}.}
\label{tab:layerselection}
\vspace{-2mm}
\end{table}
\section{Conclusion}
\label{sec:conclusion}

We proposed \method, a novel search framework for automatically configuring parameter-efficient fine-tuning (PEFT) modules of pretrained language models. \method features both a large and expressive, newly designed configuration \emph{search space} and an effective \emph{search method} featuring Bayesian optimisation that discovers a Pareto-optimal set of novel PEFT configurations with promising performance-efficiency trade-offs. Empirically, we demonstrated that \method-discovered configurations transfer strongly across different GLUE and SuperGLUE tasks, outperforming various strong PEFT baselines and being competitive to full model fine-tuning. %

\section*{Limitations and Future Work}
\method search {inevitably incurs a search cost} since it requires iterative optimisation at search time. 
However, we mitigate this by (i) using a low-fidelity proxy of 1-epoch training and (ii) leveraging strong transferability by generalising from low-resource and, thus, quick-to-train tasks. While the search itself can be seen as a \textit{one-time }cost yielding a \textit{permanent} well-performing and shareable configuration for particular tasks, we plan to delve deeper into further optimising the search cost in future work. 

Furthermore, while we conduct extensive experiments on the search space that contains three existing PEFT modules as building blocks, novel PEFT modules may emerge. However, \method framework is general, so we may easily integrate these forthcoming new modules. We defer thorough investigations to future work.

\section*{Acknowledgements}
Han Zhou is supported by the UK Research and Innovation (UKRI) Frontier Research Grant EP/Y031350/1 (the UK government’s funding guarantee for ERC Advanced Grants) awarded to Anna Korhonen at the University of Cambridge. Xingchen Wan is supported by the Clarendon Scholarship at the University of Oxford. The work has been supported in part by a Royal Society University Research Fellowship (no 221137; 2022-) awarded to Ivan Vuli\'{c}, and by the UK EPSRC grant EP/T02450X/1. We thank TACL editors and anonymous reviewers for their constructive feedback that enabled us to strengthen our work.

\bibliography{anthology, custom, example_paper}

\begin{thebibliography}{58}
\expandafter\ifx\csname natexlab\endcsname\relax\def\natexlab#1{#1}\fi

\bibitem[{Ansell et~al.(2022)Ansell, Ponti, Korhonen, and
  Vuli{\'c}}]{ansell-etal-2022-composable}
Alan Ansell, Edoardo Ponti, Anna Korhonen, and Ivan Vuli{\'c}. 2022.
\newblock \href {https://doi.org/10.18653/v1/2022.acl-long.125} {Composable
  sparse fine-tuning for cross-lingual transfer}.
\newblock In \emph{Proceedings of the 60th Annual Meeting of the Association
  for Computational Linguistics (Volume 1: Long Papers)}, pages 1778--1796,
  Dublin, Ireland. Association for Computational Linguistics.

\bibitem[{Balandat et~al.(2020)Balandat, Karrer, Jiang, Daulton, Letham,
  Wilson, and Bakshy}]{balandat2020botorch}
Maximilian Balandat, Brian Karrer, Daniel Jiang, Samuel Daulton, Ben Letham,
  Andrew~G Wilson, and Eytan Bakshy. 2020.
\newblock \href
  {https://proceedings.neurips.cc/paper/2020/hash/f5b1b89d98b7286673128a5fb112cb9a-Abstract.html}
  {Botorch: A framework for efficient monte-carlo bayesian optimization}.
\newblock \emph{Advances in neural information processing systems},
  33:21524--21538.

\bibitem[{Ben~Zaken et~al.(2022)Ben~Zaken, Goldberg, and
  Ravfogel}]{ben-zaken-etal-2022-bitfit}
Elad Ben~Zaken, Yoav Goldberg, and Shauli Ravfogel. 2022.
\newblock \href {https://doi.org/10.18653/v1/2022.acl-short.1} {{B}it{F}it:
  Simple parameter-efficient fine-tuning for transformer-based masked
  language-models}.
\newblock In \emph{Proceedings of the 60th Annual Meeting of the Association
  for Computational Linguistics (Volume 2: Short Papers)}, pages 1--9, Dublin,
  Ireland. Association for Computational Linguistics.

\bibitem[{Brown et~al.(2020)Brown, Mann, Ryder, Subbiah, Kaplan, Dhariwal,
  Neelakantan, Shyam, Sastry, Askell, Agarwal, Herbert{-}Voss, Krueger,
  Henighan, Child, Ramesh, Ziegler, Wu, Winter, Hesse, Chen, Sigler, Litwin,
  Gray, Chess, Clark, Berner, McCandlish, Radford, Sutskever, and
  Amodei}]{DBLP:conf/nips/BrownMRSKDNSSAA20}
Tom~B. Brown, Benjamin Mann, Nick Ryder, Melanie Subbiah, Jared Kaplan,
  Prafulla Dhariwal, Arvind Neelakantan, Pranav Shyam, Girish Sastry, Amanda
  Askell, Sandhini Agarwal, Ariel Herbert{-}Voss, Gretchen Krueger, Tom
  Henighan, Rewon Child, Aditya Ramesh, Daniel~M. Ziegler, Jeffrey Wu, Clemens
  Winter, Christopher Hesse, Mark Chen, Eric Sigler, Mateusz Litwin, Scott
  Gray, Benjamin Chess, Jack Clark, Christopher Berner, Sam McCandlish, Alec
  Radford, Ilya Sutskever, and Dario Amodei. 2020.
\newblock \href
  {https://proceedings.neurips.cc/paper/2020/hash/1457c0d6bfcb4967418bfb8ac142f64a-Abstract.html}
  {Language models are few-shot learners}.
\newblock In \emph{Advances in Neural Information Processing Systems 33: Annual
  Conference on Neural Information Processing Systems 2020, NeurIPS 2020,
  December 6-12, 2020, virtual}.

\bibitem[{Chen et~al.(2022{\natexlab{a}})Chen, Liu, Meng, and
  Liang}]{Chen:2022emnlp}
Guanzheng Chen, Fangyu Liu, Zaiqiao Meng, and Shangsong Liang.
  2022{\natexlab{a}}.
\newblock \href {https://doi.org/10.18653/v1/2022.emnlp-main.168} {Revisiting
  parameter-efficient tuning: Are we really there yet?}
\newblock In \emph{Proceedings of the 2022 Conference on Empirical Methods in
  Natural Language Processing}, pages 2612--2626, Abu Dhabi, United Arab
  Emirates. Association for Computational Linguistics.

\bibitem[{Chen et~al.(2023)Chen, Zhang, Shi, Li, Smola, and
  Yang}]{chen2023parameterefficient}
Jiaao Chen, Aston Zhang, Xingjian Shi, Mu~Li, Alex Smola, and Diyi Yang. 2023.
\newblock \href {https://openreview.net/forum?id=XSRSWxyJIC}
  {Parameter-efficient fine-tuning design spaces}.
\newblock In \emph{The Eleventh International Conference on Learning
  Representations}.

\bibitem[{Chen et~al.(2022{\natexlab{b}})Chen, Ge, Tong, Wang, Song, Wang, and
  Luo}]{https://doi.org/10.48550/arxiv.2205.13535}
Shoufa Chen, Chongjian Ge, Zhan Tong, Jiangliu Wang, Yibing Song, Jue Wang, and
  Ping Luo. 2022{\natexlab{b}}.
\newblock \href
  {http://papers.nips.cc/paper\_files/paper/2022/hash/69e2f49ab0837b71b0e0cb7c555990f8-Abstract-Conference.html}
  {Adaptformer: Adapting vision transformers for scalable visual recognition}.
\newblock In \emph{Advances in Neural Information Processing Systems 35: Annual
  Conference on Neural Information Processing Systems 2022, NeurIPS 2022, New
  Orleans, LA, USA, November 28 - December 9, 2022}.

\bibitem[{Daulton et~al.(2021)Daulton, Balandat, and
  Bakshy}]{daulton2021parallel}
Samuel Daulton, Maximilian Balandat, and Eytan Bakshy. 2021.
\newblock \href
  {https://proceedings.neurips.cc/paper/2021/hash/11704817e347269b7254e744b5e22dac-Abstract.html}
  {Parallel bayesian optimization of multiple noisy objectives with expected
  hypervolume improvement}.
\newblock In \emph{Advances in Neural Information Processing Systems 34: Annual
  Conference on Neural Information Processing Systems 2021, NeurIPS 2021,
  December 6-14, 2021, virtual}, pages 2187--2200.

\bibitem[{Devlin et~al.(2019)Devlin, Chang, Lee, and
  Toutanova}]{devlin-etal-2019-bert}
Jacob Devlin, Ming-Wei Chang, Kenton Lee, and Kristina Toutanova. 2019.
\newblock \href {https://doi.org/10.18653/v1/N19-1423} {{BERT}: Pre-training of
  deep bidirectional transformers for language understanding}.
\newblock In \emph{Proceedings of the 2019 Conference of the North {A}merican
  Chapter of the Association for Computational Linguistics: Human Language
  Technologies, Volume 1 (Long and Short Papers)}, pages 4171--4186,
  Minneapolis, Minnesota. Association for Computational Linguistics.

\bibitem[{Dong and Yang(2020)}]{dong2020bench}
Xuanyi Dong and Yi~Yang. 2020.
\newblock \href {https://openreview.net/forum?id=HJxyZkBKDr} {Nas-bench-201:
  Extending the scope of reproducible neural architecture search}.
\newblock In \emph{8th International Conference on Learning Representations,
  {ICLR} 2020, Addis Ababa, Ethiopia, April 26-30, 2020}.

\bibitem[{Elsken et~al.(2019)Elsken, Metzen, and Hutter}]{elsken2019neural}
Thomas Elsken, Jan~Hendrik Metzen, and Frank Hutter. 2019.
\newblock \href {http://jmlr.org/papers/v20/18-598.html} {Neural architecture
  search: A survey}.
\newblock \emph{The Journal of Machine Learning Research}, 20(1):1997--2017.

\bibitem[{Eriksson et~al.(2021)Eriksson, Chuang, Daulton, Xia, Shrivastava,
  Babu, Zhao, Aly, Venkatesh, and Balandat}]{eriksson2021latency}
David Eriksson, Pierce I-Jen Chuang, Samuel Daulton, Peng Xia, Akshat
  Shrivastava, Arun Babu, Shicong Zhao, Ahmed~A Aly, Ganesh Venkatesh, and
  Maximilian Balandat. 2021.
\newblock \href {https://openreview.net/forum?id=0ciyfd4SvbI} {Latency-aware
  neural architecture search with multi-objective bayesian optimization}.
\newblock In \emph{8th ICML Workshop on Automated Machine Learning (AutoML)}.

\bibitem[{Eriksson and Jankowiak(2021)}]{eriksson2021high}
David Eriksson and Martin Jankowiak. 2021.
\newblock \href {https://proceedings.mlr.press/v161/eriksson21a.html}
  {High-dimensional bayesian optimization with sparse axis-aligned subspaces}.
\newblock In \emph{Uncertainty in Artificial Intelligence}, pages 493--503.
  PMLR.

\bibitem[{Frazier(2018)}]{frazier2018tutorial}
Peter~I. Frazier. 2018.
\newblock \href {https://doi.org/https://doi.org/10.48550/arXiv.1807.02811} {A
  tutorial on bayesian optimization}.
\newblock \emph{CoRR}, abs/1807.02811v1.

\bibitem[{Garnett(2023)}]{garnett2023bayesian}
Roman Garnett. 2023.
\newblock \href {https://doi.org/https://doi.org/10.1017/9781108348973}
  {\emph{Bayesian Optimization}}.
\newblock Cambridge University Press.

\bibitem[{Guo et~al.(2021)Guo, Rush, and Kim}]{guo-etal-2021-parameter}
Demi Guo, Alexander Rush, and Yoon Kim. 2021.
\newblock \href {https://doi.org/10.18653/v1/2021.acl-long.378}
  {Parameter-efficient transfer learning with diff pruning}.
\newblock In \emph{Proceedings of the 59th Annual Meeting of the Association
  for Computational Linguistics and the 11th International Joint Conference on
  Natural Language Processing (Volume 1: Long Papers)}, pages 4884--4896,
  Online. Association for Computational Linguistics.

\bibitem[{He et~al.(2022)He, Zhou, Ma, Berg{-}Kirkpatrick, and
  Neubig}]{DBLP:conf/iclr/HeZMBN22}
Junxian He, Chunting Zhou, Xuezhe Ma, Taylor Berg{-}Kirkpatrick, and Graham
  Neubig. 2022.
\newblock \href {https://openreview.net/forum?id=0RDcd5Axok} {Towards a unified
  view of parameter-efficient transfer learning}.
\newblock In \emph{The Tenth International Conference on Learning
  Representations, {ICLR} 2022, Virtual Event, April 25-29, 2022}.

\bibitem[{Houlsby et~al.(2019)Houlsby, Giurgiu, Jastrzebski, Morrone,
  de~Laroussilhe, Gesmundo, Attariyan, and
  Gelly}]{DBLP:conf/icml/HoulsbyGJMLGAG19}
Neil Houlsby, Andrei Giurgiu, Stanislaw Jastrzebski, Bruna Morrone, Quentin
  de~Laroussilhe, Andrea Gesmundo, Mona Attariyan, and Sylvain Gelly. 2019.
\newblock \href {http://proceedings.mlr.press/v97/houlsby19a.html}
  {Parameter-efficient transfer learning for {NLP}}.
\newblock In \emph{Proceedings of the 36th International Conference on Machine
  Learning, {ICML} 2019, 9-15 June 2019, Long Beach, California, {USA}}, pages
  2790--2799.

\bibitem[{Hu et~al.(2022{\natexlab{a}})Hu, yelong shen, Wallis, Allen-Zhu, Li,
  Wang, Wang, and Chen}]{DBLP:conf/iclr/HuSWALWWC22}
Edward~J Hu, yelong shen, Phillip Wallis, Zeyuan Allen-Zhu, Yuanzhi Li, Shean
  Wang, Lu~Wang, and Weizhu Chen. 2022{\natexlab{a}}.
\newblock \href {https://openreview.net/forum?id=nZeVKeeFYf9} {Lo{RA}: Low-rank
  adaptation of large language models}.
\newblock In \emph{International Conference on Learning Representations}.

\bibitem[{Hu et~al.(2022{\natexlab{b}})Hu, Zhang, Ding, Wang, Wang, Liu, and
  Sun}]{DBLP:journals/corr/abs-2206-07382}
Shengding Hu, Zhen Zhang, Ning Ding, Yadao Wang, Yasheng Wang, Zhiyuan Liu, and
  Maosong Sun. 2022{\natexlab{b}}.
\newblock \href {https://openreview.net/forum?id=oOte_397Q4P} {Sparse structure
  search for delta tuning}.
\newblock In \emph{Advances in Neural Information Processing Systems}.

\bibitem[{Izquierdo et~al.(2021)Izquierdo, Guerrero-Viu, Hauns, Miotto,
  Schrodi, Biedenkapp, Elsken, Deng, Lindauer, and Hutter}]{guerrero2021bag}
Sergio Izquierdo, Julia Guerrero-Viu, Sven Hauns, Guilherme Miotto, Simon
  Schrodi, Andr{\'e} Biedenkapp, Thomas Elsken, Difan Deng, Marius Lindauer,
  and Frank Hutter. 2021.
\newblock \href {https://arxiv.org/abs/2105.01015} {Bag of baselines for
  multi-objective joint neural architecture search and hyperparameter
  optimization}.
\newblock In \emph{8th ICML Workshop on Automated Machine Learning (AutoML)}.

\bibitem[{Lester et~al.(2021)Lester, Al-Rfou, and
  Constant}]{lester-etal-2021-power}
Brian Lester, Rami Al-Rfou, and Noah Constant. 2021.
\newblock \href {https://doi.org/10.18653/v1/2021.emnlp-main.243} {The power of
  scale for parameter-efficient prompt tuning}.
\newblock In \emph{Proceedings of the 2021 Conference on Empirical Methods in
  Natural Language Processing}, pages 3045--3059, Online and Punta Cana,
  Dominican Republic. Association for Computational Linguistics.

\bibitem[{Li and Talwalkar(2019)}]{li2020random}
Liam Li and Ameet Talwalkar. 2019.
\newblock \href {http://proceedings.mlr.press/v115/li20c.html} {Random search
  and reproducibility for neural architecture search}.
\newblock In \emph{Proceedings of the Thirty-Fifth Conference on Uncertainty in
  Artificial Intelligence, {UAI} 2019, Tel Aviv, Israel, July 22-25, 2019},
  pages 367--377.

\bibitem[{Li and Liang(2021)}]{li-liang-2021-prefix}
Xiang~Lisa Li and Percy Liang. 2021.
\newblock \href {https://doi.org/10.18653/v1/2021.acl-long.353} {Prefix-tuning:
  Optimizing continuous prompts for generation}.
\newblock In \emph{Proceedings of the 59th Annual Meeting of the Association
  for Computational Linguistics and the 11th International Joint Conference on
  Natural Language Processing (Volume 1: Long Papers)}, pages 4582--4597,
  Online. Association for Computational Linguistics.

\bibitem[{Liu et~al.(2019{\natexlab{a}})Liu, Simonyan, and Yang}]{liu2018darts}
Hanxiao Liu, Karen Simonyan, and Yiming Yang. 2019{\natexlab{a}}.
\newblock \href {https://openreview.net/forum?id=S1eYHoC5FX} {{DARTS:}
  differentiable architecture search}.
\newblock In \emph{7th International Conference on Learning Representations,
  {ICLR} 2019, New Orleans, LA, USA, May 6-9, 2019}.

\bibitem[{Liu et~al.(2022)Liu, Tam, Mohammed, Mohta, Huang, Bansal, and
  Raffel}]{DBLP:journals/corr/abs-2205-05638}
Haokun Liu, Derek Tam, Muqeeth Mohammed, Jay Mohta, Tenghao Huang, Mohit
  Bansal, and Colin Raffel. 2022.
\newblock \href {https://openreview.net/forum?id=rBCvMG-JsPd} {Few-shot
  parameter-efficient fine-tuning is better and cheaper than in-context
  learning}.
\newblock In \emph{Advances in Neural Information Processing Systems}.

\bibitem[{Liu et~al.(2019{\natexlab{b}})Liu, Ott, Goyal, Du, Joshi, Chen, Levy,
  Lewis, Zettlemoyer, and Stoyanov}]{DBLP:journals/corr/abs-1907-11692}
Yinhan Liu, Myle Ott, Naman Goyal, Jingfei Du, Mandar Joshi, Danqi Chen, Omer
  Levy, Mike Lewis, Luke Zettlemoyer, and Veselin Stoyanov. 2019{\natexlab{b}}.
\newblock \href {http://arxiv.org/abs/1907.11692} {Roberta: {A} robustly
  optimized {BERT} pretraining approach}.
\newblock \emph{CoRR}, abs/1907.11692v1.

\bibitem[{Mahabadi et~al.(2021)Mahabadi, Henderson, and
  Ruder}]{DBLP:conf/nips/MahabadiHR21}
Rabeeh~Karimi Mahabadi, James Henderson, and Sebastian Ruder. 2021.
\newblock \href
  {https://proceedings.neurips.cc/paper/2021/hash/081be9fdff07f3bc808f935906ef70c0-Abstract.html}
  {Compacter: Efficient low-rank hypercomplex adapter layers}.
\newblock In \emph{Advances in Neural Information Processing Systems 34: Annual
  Conference on Neural Information Processing Systems 2021, NeurIPS 2021,
  December 6-14, 2021, virtual}, pages 1022--1035.

\bibitem[{Mao et~al.(2022)Mao, Mathias, Hou, Almahairi, Ma, Han, Yih, and
  Khabsa}]{mao-etal-2022-unipelt}
Yuning Mao, Lambert Mathias, Rui Hou, Amjad Almahairi, Hao Ma, Jiawei Han,
  Scott Yih, and Madian Khabsa. 2022.
\newblock \href {https://doi.org/10.18653/v1/2022.acl-long.433} {{U}ni{PELT}: A
  unified framework for parameter-efficient language model tuning}.
\newblock In \emph{Proceedings of the 60th Annual Meeting of the Association
  for Computational Linguistics (Volume 1: Long Papers)}, pages 6253--6264,
  Dublin, Ireland. Association for Computational Linguistics.

\bibitem[{Moosavi et~al.(2022)Moosavi, Delfosse, Kersting, and
  Gurevych}]{moosavi-etal-2022-adaptable}
Nafise Moosavi, Quentin Delfosse, Kristian Kersting, and Iryna Gurevych. 2022.
\newblock \href {https://doi.org/10.18653/v1/2022.naacl-main.274} {Adaptable
  adapters}.
\newblock In \emph{Proceedings of the 2022 Conference of the North American
  Chapter of the Association for Computational Linguistics: Human Language
  Technologies}, pages 3742--3753, Seattle, United States. Association for
  Computational Linguistics.

\bibitem[{Pfeiffer et~al.(2022)Pfeiffer, Goyal, Lin, Li, Cross, Riedel, and
  Artetxe}]{pfeiffer-etal-2022-lifting}
Jonas Pfeiffer, Naman Goyal, Xi~Lin, Xian Li, James Cross, Sebastian Riedel,
  and Mikel Artetxe. 2022.
\newblock \href {https://doi.org/10.18653/v1/2022.naacl-main.255} {Lifting the
  curse of multilinguality by pre-training modular transformers}.
\newblock In \emph{Proceedings of the 2022 Conference of the North American
  Chapter of the Association for Computational Linguistics: Human Language
  Technologies}, pages 3479--3495, Seattle, United States. Association for
  Computational Linguistics.

\bibitem[{Pfeiffer et~al.(2020{\natexlab{a}})Pfeiffer, R{\"u}ckl{\'e}, Poth,
  Kamath, Vuli{\'c}, Ruder, Cho, and Gurevych}]{pfeiffer-etal-2020-adapterhub}
Jonas Pfeiffer, Andreas R{\"u}ckl{\'e}, Clifton Poth, Aishwarya Kamath, Ivan
  Vuli{\'c}, Sebastian Ruder, Kyunghyun Cho, and Iryna Gurevych.
  2020{\natexlab{a}}.
\newblock \href {https://doi.org/10.18653/v1/2020.emnlp-demos.7}
  {{A}dapter{H}ub: A framework for adapting transformers}.
\newblock In \emph{Proceedings of the 2020 Conference on Empirical Methods in
  Natural Language Processing: System Demonstrations}, pages 46--54, Online.
  Association for Computational Linguistics.

\bibitem[{Pfeiffer et~al.(2023)Pfeiffer, Ruder, Vuli{\'c}, and
  Ponti}]{Pfeiffer:2023survey}
Jonas Pfeiffer, Sebastian Ruder, Ivan Vuli{\'c}, and Edoardo Ponti. 2023.
\newblock \href {https://doi.org/10.48550/arXiv.2302.11529} {Modular deep
  learning}.
\newblock \emph{Transactions on Machine Learning Research}.
\newblock Survey Certification.

\bibitem[{Pfeiffer et~al.(2020{\natexlab{b}})Pfeiffer, Vuli{\'c}, Gurevych, and
  Ruder}]{pfeiffer-etal-2020-mad}
Jonas Pfeiffer, Ivan Vuli{\'c}, Iryna Gurevych, and Sebastian Ruder.
  2020{\natexlab{b}}.
\newblock \href {https://doi.org/10.18653/v1/2020.emnlp-main.617} {{MAD-X}:
  {A}n {A}dapter-{B}ased {F}ramework for {M}ulti-{T}ask {C}ross-{L}ingual
  {T}ransfer}.
\newblock In \emph{Proceedings of the 2020 Conference on Empirical Methods in
  Natural Language Processing (EMNLP)}, pages 7654--7673, Online. Association
  for Computational Linguistics.

\bibitem[{Raffel et~al.(2020{\natexlab{a}})Raffel, Shazeer, Roberts, Lee,
  Narang, Matena, Zhou, Li, and Liu}]{DBLP:journals/jmlr/RaffelSRLNMZLL20}
Colin Raffel, Noam Shazeer, Adam Roberts, Katherine Lee, Sharan Narang, Michael
  Matena, Yanqi Zhou, Wei Li, and Peter~J. Liu. 2020{\natexlab{a}}.
\newblock \href {http://jmlr.org/papers/v21/20-074.html} {Exploring the limits
  of transfer learning with a unified text-to-text transformer}.
\newblock \emph{J. Mach. Learn. Res.}, 21:140:1--140:67.

\bibitem[{Raffel et~al.(2020{\natexlab{b}})Raffel, Shazeer, Roberts, Lee,
  Narang, Matena, Zhou, Li, and Liu}]{raffel2020t5}
Colin Raffel, Noam Shazeer, Adam Roberts, Katherine Lee, Sharan Narang, Michael
  Matena, Yanqi Zhou, Wei Li, and Peter~J. Liu. 2020{\natexlab{b}}.
\newblock \href {http://jmlr.org/papers/v21/20-074.html} {Exploring the limits
  of transfer learning with a unified text-to-text transformer}.
\newblock \emph{J. Mach. Learn. Res.}, 21:140:1--140:67.

\bibitem[{Ren et~al.(2021)Ren, Xiao, Chang, Huang, Li, Chen, and
  Wang}]{ren2021comprehensive}
Pengzhen Ren, Yun Xiao, Xiaojun Chang, Po-Yao Huang, Zhihui Li, Xiaojiang Chen,
  and Xin Wang. 2021.
\newblock \href {https://doi.org/10.1145/3447582} {A comprehensive survey of
  neural architecture search: Challenges and solutions}.
\newblock \emph{ACM Computing Surveys (CSUR)}, 54(4):1--34.

\bibitem[{Ru et~al.(2021)Ru, Wan, Dong, and Osborne}]{ru2020interpretable}
Bin~Xin Ru, Xingchen Wan, Xiaowen Dong, and Michael~A. Osborne. 2021.
\newblock \href {https://openreview.net/forum?id=j9Rv7qdXjd} {Interpretable
  neural architecture search via bayesian optimisation with weisfeiler-lehman
  kernels}.
\newblock In \emph{9th International Conference on Learning Representations,
  {ICLR} 2021, Virtual Event, Austria, May 3-7, 2021}.

\bibitem[{Ru et~al.(2020)Ru, Esperan{\c{c}}a, and Carlucci}]{ru2020neural}
Robin Ru, Pedro~M. Esperan{\c{c}}a, and Fabio~Maria Carlucci. 2020.
\newblock \href
  {https://proceedings.neurips.cc/paper/2020/hash/8c53d30ad023ce50140181f713059ddf-Abstract.html}
  {Neural architecture generator optimization}.
\newblock In \emph{Advances in Neural Information Processing Systems 33: Annual
  Conference on Neural Information Processing Systems 2020, NeurIPS 2020,
  December 6-12, 2020, virtual}.

\bibitem[{R{\"u}ckl{\'e} et~al.(2021)R{\"u}ckl{\'e}, Geigle, Glockner, Beck,
  Pfeiffer, Reimers, and Gurevych}]{ruckle-etal-2021-adapterdrop}
Andreas R{\"u}ckl{\'e}, Gregor Geigle, Max Glockner, Tilman Beck, Jonas
  Pfeiffer, Nils Reimers, and Iryna Gurevych. 2021.
\newblock \href {https://doi.org/10.18653/v1/2021.emnlp-main.626}
  {{AdapterDrop}: {O}n the efficiency of adapters in transformers}.
\newblock In \emph{Proceedings of the 2021 Conference on Empirical Methods in
  Natural Language Processing}, pages 7930--7946, Online and Punta Cana,
  Dominican Republic. Association for Computational Linguistics.

\bibitem[{Sanh et~al.(2022)Sanh, Webson, Raffel, Bach, Sutawika, Alyafeai,
  Chaffin, Stiegler, Raja, Dey, Bari, Xu, Thakker, Sharma, Szczechla, Kim,
  Chhablani, Nayak, Datta, Chang, Jiang, Wang, Manica, Shen, Yong, Pandey,
  Bawden, Wang, Neeraj, Rozen, Sharma, Santilli, F{\'{e}}vry, Fries, Teehan,
  Scao, Biderman, Gao, Wolf, and Rush}]{DBLP:conf/iclr/SanhWRBSACSRDBX22}
Victor Sanh, Albert Webson, Colin Raffel, Stephen Bach, Lintang Sutawika, Zaid
  Alyafeai, Antoine Chaffin, Arnaud Stiegler, Arun Raja, Manan Dey, M~Saiful
  Bari, Canwen Xu, Urmish Thakker, Shanya~Sharma Sharma, Eliza Szczechla,
  Taewoon Kim, Gunjan Chhablani, Nihal~V. Nayak, Debajyoti Datta, Jonathan
  Chang, Mike~Tian{-}Jian Jiang, Han Wang, Matteo Manica, Sheng Shen, Zheng~Xin
  Yong, Harshit Pandey, Rachel Bawden, Thomas Wang, Trishala Neeraj, Jos Rozen,
  Abheesht Sharma, Andrea Santilli, Thibault F{\'{e}}vry, Jason~Alan Fries,
  Ryan Teehan, Teven~Le Scao, Stella Biderman, Leo Gao, Thomas Wolf, and
  Alexander~M. Rush. 2022.
\newblock \href {https://openreview.net/forum?id=9Vrb9D0WI4} {Multitask
  prompted training enables zero-shot task generalization}.
\newblock In \emph{The Tenth International Conference on Learning
  Representations, {ICLR} 2022, Virtual Event, April 25-29, 2022}.

\bibitem[{Sung et~al.(2021)Sung, Nair, and Raffel}]{DBLP:conf/nips/SungNR21}
Yi{-}Lin Sung, Varun Nair, and Colin Raffel. 2021.
\newblock \href
  {https://proceedings.neurips.cc/paper/2021/hash/cb2653f548f8709598e8b5156738cc51-Abstract.html}
  {Training neural networks with fixed sparse masks}.
\newblock In \emph{Advances in Neural Information Processing Systems 34: Annual
  Conference on Neural Information Processing Systems 2021, NeurIPS 2021,
  December 6-14, 2021, virtual}, pages 24193--24205.

\bibitem[{Tenney et~al.(2019)Tenney, Das, and Pavlick}]{tenney-etal-2019-bert}
Ian Tenney, Dipanjan Das, and Ellie Pavlick. 2019.
\newblock \href {https://doi.org/10.18653/v1/P19-1452} {{BERT} rediscovers the
  classical {NLP} pipeline}.
\newblock In \emph{Proceedings of the 57th Annual Meeting of the Association
  for Computational Linguistics}, pages 4593--4601, Florence, Italy.
  Association for Computational Linguistics.

\bibitem[{Valipour et~al.(2023)Valipour, Rezagholizadeh, Kobyzev, and
  Ghodsi}]{DBLP:journals/corr/abs-2210-07558}
Mojtaba Valipour, Mehdi Rezagholizadeh, Ivan Kobyzev, and Ali Ghodsi. 2023.
\newblock \href {https://doi.org/10.18653/v1/2023.eacl-main.239} {{D}y{L}o{RA}:
  Parameter-efficient tuning of pre-trained models using dynamic search-free
  low-rank adaptation}.
\newblock In \emph{Proceedings of the 17th Conference of the European Chapter
  of the Association for Computational Linguistics}, pages 3274--3287,
  Dubrovnik, Croatia. Association for Computational Linguistics.

\bibitem[{Vuli{\'c} et~al.(2020)Vuli{\'c}, Ponti, Litschko, Glava{\v{s}}, and
  Korhonen}]{vulic-etal-2020-probing}
Ivan Vuli{\'c}, Edoardo~Maria Ponti, Robert Litschko, Goran Glava{\v{s}}, and
  Anna Korhonen. 2020.
\newblock \href {https://doi.org/10.18653/v1/2020.emnlp-main.586} {Probing
  pretrained language models for lexical semantics}.
\newblock In \emph{Proceedings of the 2020 Conference on Empirical Methods in
  Natural Language Processing (EMNLP)}, pages 7222--7240, Online. Association
  for Computational Linguistics.

\bibitem[{Wan et~al.(2021)Wan, Nguyen, Ha, Ru, Lu, and Osborne}]{wan2021think}
Xingchen Wan, Vu~Nguyen, Huong Ha, Binxin Ru, Cong Lu, and Michael~A Osborne.
  2021.
\newblock \href {http://proceedings.mlr.press/v139/wan21b.html} {Think global
  and act local: Bayesian optimisation over high-dimensional categorical and
  mixed search spaces}.
\newblock In \emph{International Conference on Machine Learning}, pages
  10663--10674. PMLR.

\bibitem[{Wan et~al.(2022)Wan, Ru, Esperan{\c{c}}a, and
  Li}]{DBLP:conf/iclr/WanREL22}
Xingchen Wan, Binxin Ru, Pedro~M. Esperan{\c{c}}a, and Zhenguo Li. 2022.
\newblock \href {https://openreview.net/forum?id=rFJWoYoxrDB} {On redundancy
  and diversity in cell-based neural architecture search}.
\newblock In \emph{The Tenth International Conference on Learning
  Representations, {ICLR} 2022, Virtual Event, April 25-29, 2022}.
  OpenReview.net.

\bibitem[{Wang et~al.(2019)Wang, Pruksachatkun, Nangia, Singh, Michael, Hill,
  Levy, and Bowman}]{DBLP:conf/nips/WangPNSMHLB19}
Alex Wang, Yada Pruksachatkun, Nikita Nangia, Amanpreet Singh, Julian Michael,
  Felix Hill, Omer Levy, and Samuel~R. Bowman. 2019.
\newblock \href
  {https://proceedings.neurips.cc/paper/2019/hash/4496bf24afe7fab6f046bf4923da8de6-Abstract.html}
  {Superglue: {A} stickier benchmark for general-purpose language understanding
  systems}.
\newblock In \emph{Advances in Neural Information Processing Systems 32: Annual
  Conference on Neural Information Processing Systems 2019, NeurIPS 2019,
  December 8-14, 2019, Vancouver, BC, Canada}, pages 3261--3275.

\bibitem[{Wang et~al.(2018)Wang, Singh, Michael, Hill, Levy, and
  Bowman}]{wang-etal-2018-glue}
Alex Wang, Amanpreet Singh, Julian Michael, Felix Hill, Omer Levy, and Samuel
  Bowman. 2018.
\newblock \href {https://doi.org/10.18653/v1/W18-5446} {{GLUE}: A multi-task
  benchmark and analysis platform for natural language understanding}.
\newblock In \emph{Proceedings of the 2018 {EMNLP} Workshop {B}lackbox{NLP}:
  Analyzing and Interpreting Neural Networks for {NLP}}, pages 353--355,
  Brussels, Belgium. Association for Computational Linguistics.

\bibitem[{Wang et~al.(2022)Wang, Agarwal, Mukherjee, Liu, Gao, Awadallah, and
  Gao}]{https://doi.org/10.48550/arxiv.2205.12410}
Yaqing Wang, Sahaj Agarwal, Subhabrata Mukherjee, Xiaodong Liu, Jing Gao,
  Ahmed~Hassan Awadallah, and Jianfeng Gao. 2022.
\newblock \href {https://doi.org/10.18653/v1/2022.emnlp-main.388} {{A}da{M}ix:
  Mixture-of-adaptations for parameter-efficient model tuning}.
\newblock In \emph{Proceedings of the 2022 Conference on Empirical Methods in
  Natural Language Processing}, pages 5744--5760, Abu Dhabi, United Arab
  Emirates. Association for Computational Linguistics.

\bibitem[{White et~al.(2021)White, Neiswanger, and Savani}]{white2021bananas}
Colin White, Willie Neiswanger, and Yash Savani. 2021.
\newblock \href {https://ojs.aaai.org/index.php/AAAI/article/view/17233}
  {{BANANAS:} bayesian optimization with neural architectures for neural
  architecture search}.
\newblock In \emph{Thirty-Fifth {AAAI} Conference on Artificial Intelligence,
  {AAAI} 2021, Thirty-Third Conference on Innovative Applications of Artificial
  Intelligence, {IAAI} 2021, The Eleventh Symposium on Educational Advances in
  Artificial Intelligence, {EAAI} 2021, Virtual Event, February 2-9, 2021},
  pages 10293--10301.

\bibitem[{Xie et~al.(2019)Xie, Kirillov, Girshick, and He}]{xie2019exploring}
Saining Xie, Alexander Kirillov, Ross~B. Girshick, and Kaiming He. 2019.
\newblock \href {https://doi.org/10.1109/ICCV.2019.00137} {Exploring randomly
  wired neural networks for image recognition}.
\newblock In \emph{2019 {IEEE/CVF} International Conference on Computer Vision,
  {ICCV} 2019, Seoul, Korea (South), October 27 - November 2, 2019}, pages
  1284--1293. {IEEE}.

\bibitem[{Yang et~al.(2020)Yang, Esperan{\c{c}}a, and
  Carlucci}]{yang2019evaluation}
Antoine Yang, Pedro~M. Esperan{\c{c}}a, and Fabio~Maria Carlucci. 2020.
\newblock \href {https://openreview.net/forum?id=HygrdpVKvr} {{NAS} evaluation
  is frustratingly hard}.
\newblock In \emph{8th International Conference on Learning Representations,
  {ICLR} 2020, Addis Ababa, Ethiopia, April 26-30, 2020}.

\bibitem[{Zhang et~al.(2023)Zhang, Chen, Bukharin, He, Cheng, Chen, and
  Zhao}]{zhang2023adaptive}
Qingru Zhang, Minshuo Chen, Alexander Bukharin, Pengcheng He, Yu~Cheng, Weizhu
  Chen, and Tuo Zhao. 2023.
\newblock \href {https://openreview.net/forum?id=lq62uWRJjiY} {Adaptive budget
  allocation for parameter-efficient fine-tuning}.
\newblock In \emph{The Eleventh International Conference on Learning
  Representations}.

\bibitem[{Zhao et~al.(2020)Zhao, Lin, Mi, Jaggi, and
  Sch{\"u}tze}]{zhao-etal-2020-masking}
Mengjie Zhao, Tao Lin, Fei Mi, Martin Jaggi, and Hinrich Sch{\"u}tze. 2020.
\newblock \href {https://doi.org/10.18653/v1/2020.emnlp-main.174} {Masking as
  an efficient alternative to finetuning for pretrained language models}.
\newblock In \emph{Proceedings of the 2020 Conference on Empirical Methods in
  Natural Language Processing (EMNLP)}, pages 2226--2241, Online. Association
  for Computational Linguistics.

\bibitem[{Zhou et~al.(2023)Zhou, Wan, Vuli{\'c}, and
  Korhonen}]{zhou2023survival}
Han Zhou, Xingchen Wan, Ivan Vuli{\'c}, and Anna Korhonen. 2023.
\newblock \href {https://doi.org/10.18653/v1/2023.findings-emnlp.870} {Survival
  of the most influential prompts: Efficient black-box prompt search via
  clustering and pruning}.
\newblock In \emph{Findings of the Association for Computational Linguistics:
  EMNLP 2023}, pages 13064--13077, Singapore. Association for Computational
  Linguistics.

\bibitem[{Zitzler and Thiele(1998)}]{zitzler1998multiobjective}
Eckart Zitzler and Lothar Thiele. 1998.
\newblock \href {https://doi.org/10.1007/BFB0056872} {Multiobjective
  optimization using evolutionary algorithms - {A} comparative case study}.
\newblock In \emph{Parallel Problem Solving from Nature - {PPSN} V, 5th
  International Conference, Amsterdam, The Netherlands, September 27-30, 1998,
  Proceedings}, volume 1498 of \emph{Lecture Notes in Computer Science}, pages
  292--304. Springer.

\bibitem[{Zoph and Le(2017)}]{zoph2016neural}
Barret Zoph and Quoc~V. Le. 2017.
\newblock \href {https://openreview.net/forum?id=r1Ue8Hcxg} {Neural
  architecture search with reinforcement learning}.
\newblock In \emph{5th International Conference on Learning Representations,
  {ICLR} 2017, Toulon, France, April 24-26, 2017, Conference Track
  Proceedings}.

\end{thebibliography}
\bibliographystyle{acl_natbib}
\end{document}